\documentclass{article}
\usepackage{iclr2025_conference,times}

\usepackage[utf8]{inputenc} 
\usepackage{CJKutf8}
\usepackage{newunicodechar}
\newunicodechar{▁}{\textunderscore}

\usepackage{amsmath,amsfonts,bm}

\def\eqref#1{equation~\ref{#1}}

\def\1{\bm{1}}

\DeclareMathAlphabet{\mathsfit}{\encodingdefault}{\sfdefault}{m}{sl}
\SetMathAlphabet{\mathsfit}{bold}{\encodingdefault}{\sfdefault}{bx}{n}

\usepackage{hyperref}
\usepackage{url}
\usepackage{booktabs}
\usepackage{multirow} 
\usepackage{array}
\usepackage{graphicx}
\usepackage{subcaption}
\usepackage{caption}
\usepackage{wasysym}
\usepackage{marvosym}
\usepackage{algorithm}
\usepackage[noend]{algpseudocode}
\usepackage{enumitem}

\usepackage{amsmath,amsfonts,amssymb,amsthm}

\title{Jet Expansions of Residual Computation}

\author{
Yihong Chen$^{\text{\Aries}}$, Xiangxiang Xu$^{\text{\Aquarius}}$, Yao Lu$^{\text{\Aries}}$, Pontus Stenetorp$^{\text{\Aries}}$, Luca Franceschi$^{\text{\Libra}}$ \\
$^{\text{\Aries}}$UCL Centre for Artificial Intelligence, London, UK \\
$^{\text{\Aquarius}}$MIT, EECS, USA \quad
$^{\text{\Libra}}$Amazon Web Services, Berlin, Germany \\
\texttt{\{yihong.chen, p.stenetorp, yao.lu\}@cs.ucl.ac.uk}, 
\\
\texttt{xuxx@mit.edu}, \texttt{franuluc@amazon.de}
}

\usepackage{cleveref}
\newtheorem{remark}{Remark}
\newtheorem{lemma}{Lemma}
\newtheorem{proposition}{Proposition}

\iclrfinalcopy %
\begin{document}

\newcommand{\centerptr}[1]{\textcolor{red}{#1}} %
\newcommand{\variateptr}[1]{\textcolor{blue}{#1}} %
\newcommand{\cx}{\centerptr{x}}
\newcommand{\cxi}{\centerptr{x_i}}
\newcommand{\vy}{\variateptr{y}}

\newcommand{\rd}{\mathbb{R}^d}
\newcommand{\jet}[1][k]{\mathrm{J}^{#1}}
\newcommand{\decompose}{\texttt{jet\_expand}}

\maketitle

\begin{abstract}
We introduce a framework for expanding residual computational graphs using \textit{jets}, operators that generalize truncated Taylor series.
Our method provides a systematic approach to disentangle contributions of different computational paths to model predictions.
In contrast to existing techniques such as distillation, probing, or early decoding, our expansions rely solely on the model itself and requires no data, training, or sampling from the model.
We demonstrate how our framework grounds and subsumes \textit{logit lens},
reveals a (super-)exponential path structure in the recursive residual depth and opens up several applications. 
These include sketching a transformer large language model with $n$-gram statistics extracted from its computations, and indexing the models' levels of toxicity knowledge.
Our approach enables \textit{data-free} analysis of residual computation for model interpretability, development, and evaluation. The project website can be found \href{https://yihong-chen.github.io/jet_expand/}{here}.
\end{abstract}

\section{Introduction}
Machine learning models, particularly large-scale foundation models, have become increasingly prevalent and impactful across a wide range of domains \citep{wei2021finetuned, bommasani2023foundation, touvron2023llama}.
While delivering strong results, their black-box nature has led to the development of techniques to assess their behavior and gain insights into their internal mechanisms.
In this space, mechanistic interpretability~(MI)~\citep[see e.g.][ for recent surverys]{bereska2024mechanistic,ferrando2024primer} 
has emerged as an alternative to more classic local attribution methods such as SHAP~\citep{lundberg2017unified} or integrated gradient~\citep{sundararajan2017axiomatic}.
Contrary to these methods, which seeks to trace output behavior back to the network input,  MI focuses on tracing behavior back to the model itself. 
It seeks to uncover learned ``algorithms'' that are embedded in the model weights and computational structure,  with the aim of developing a global understanding of -- and, ultimately, to reverse engineer -- neural computation.

The great majority of MI work uses a hypothesis-and-dataset-driven approach~(see for example \citet{goldowsky2023localizing}), in that it first formalizes a hypothesis, then chooses or curates a dataset to probe the model, it applies techniques such as path patching~\citep{wang2022interpretability} or causal tracing~\citep{meng2022locating}, and then possibly refines the initial hypothesis.
While this approach to MI is valuable, it can limit the ability to perform open-ended exploration-driven studies aimed at uncovering global behavior and charting ``maps'' that connect  computation to behavior.
In this regard, studies such as \citet{veit2016residual} or \citet{elhage2021mathematical} focus on the intrinsic computation that is carried out by a model, offering complementary views to the hypothesis-and-dataset-driven approach.
Yet, these studies often make unrealistic assumptions of the model, making it unclear how much of the derived understanding can be transferred to real-world models and applications.

This paper contributes to this latter direction, presenting a general-purpose framework to manipulate the computational graph of a neural model with the aim of identifying individual input-to-output computational paths, which we can then further analyze to extract behavior.
Our method is based on the simple observation that we can recursively expand the computation of a network by selectively applying \textit{jet operators}~\citep{ehresmann1951}, which one can think of as the functional counterpart of truncated Taylor series. 
This process, which we call the \textit{jet expansion} of a model, gives rise to a class of equivalent functional rewritings of the original network into the sum of polynomial terms~(which we see as input-to-output functions and dub \textit{jet paths}) and non-linear remainders. 
The framework does not make particular assumptions on the input model and, as it operates in the space of functions~(rather than function evaluations), it requires no input data.
For transformer language models, we also show how specific instantiations linked to $n$-gram models make it feasible to exhaustively evaluate the jet paths over the entire input space, enabling end-to-end data-free global interpretability studies.

In this work, we focus on residual networks \citep{he2016deep} -- particularly transformers \citep{vaswani2017attention} -- and operate at the granularity of residual blocks~(e.g., self-attention or MLP blocks).
This approach simplifies our presentation, while aligning with previous literature such as \citep{veit2016residual}, and maintains practical relevance given the prevalence of residual networks for real-world applications. 
We describe several instantiations of our framework in \Cref{sec:notable}, showing how it encompasses previously proposed interpretability tools such as the logit lens \citep{nostalgebraist2021interpreting}. 
Based on these instantiations, we present an extensive set of case studies on several auto-regressive large language models~(LLMs) from varying families and sizes, including \textit{GPT}, \textit{Llama} and \textit{OLMo}.
Our case studies demonstrate jet expansion offers a suite of powerful tools -- jet lens, jet paths and jet $n$-grams -- to perform multi-scenario LLM interpretability: i) understanding the inner working of an LLM (\Cref{case:inner_working}); ii) debugging the pretraining dynamic (\Cref{case:pre}); and iii) examining fine-tuning effects~(\Cref{case:fte}), which are useful for improving transparent and responsible usages of LLMs.  
We close with a discussion about the potential directions of future research that this work opens, alongside its current limitations.

\section{Residual networks and their rewritings}

We start by reviewing the archetypal computational structure of residual networks and discuss the case of linear residual networks as a canonical example of functions that are intrinsically expanded.

\vspace{-3mm}
\paragraph{Residual networks.} 
We focus on network architectures whose main body  consists of multiple recursive residual blocks, while the input and output are managed respectively by an encoding and a decoding module.
Let $\mathcal{Z}$ be an input space (e.g., sequences of tokens), $c\in\mathbb{N}^+$ be the number of classes (e.g., a vocabulary size), $\mathcal{Y}=\mathbb{R}^c$ be a space of output logits and $d\in\mathbb{N}^+$ be a hidden dimension. 
Formally, we are concerned with functions $q:\mathcal{Z}\to\mathcal{Y}$ described as follows:
\begin{equation}
    \label{eq:q0}
    q = \upsilon \circ h_L, 
    \quad \text{where }\,
    h_L:\mathcal{Z}\to\rd, \;\; h_L = \bigcirc_{l=1}^{L} 
        \beta_l  
        \circ \eta,
\end{equation}
where $L\in\mathbb{N}^+$ is the number of residual blocks (e.g. recursive depth),  $\eta: \mathcal{Z} \to\rd$ is an input encoding module (e.g. token embedding layer), $\bigcirc$ denotes repeated functional composition, and 
\begin{align} 
    \label{eq:layer}
    \textstyle
    \beta_l &: \rd \to \rd \quad \text{for } \, l \in [L]
            &\beta_{l} & = \mathrm{id} + \gamma_l,  
            &\gamma_l:  \rd \to \rd,
    \\
    \upsilon & : \mathbb{R}^d \to \mathcal{Y}
             &\upsilon(x) & = U\,  \gamma_{L+1}(x)  
             & U  \in \mathbb{R}^{c\times d}, \, \gamma:  \rd \to \rd,
    \label{eq:unembedding}
\end{align}
are respectively residual blocks with nonlinearities $\gamma_l$'s (e.g., input-normalized causal self-attentions or MLPs), and the output decoding module~(e.g., an unembedding projection $U$ after a layer normalization $\gamma_{L+1}$); $\mathrm{id}$ is the identity map. 
We leave all parameters \textit{implicit} and assume all functions are $C^{\infty}$.
Optimized for classification (e.g., next token prediction for autoregressive language models), the function $q$ outputs unnormalized conditional probabilities (or logits) in that $\mathbb{P}_q(\text{``$z$ belongs to class $i$''}| z) = \mathrm{Softmax}[q(z)]_i$, for $z\in\mathcal{Z}$. 
In residual networks, the recursive links allow the ``storage'' of computation from all previous layers and the embedded input, leading to an accumulation of information across depths. 
This is highlighted by unrolling the computation of \Cref{eq:q0} up to a block $l\in[L]$, setting $h_0=\eta$:
\begin{equation}
\label{eq:unrolling}
\textstyle
h_l = \bigcirc_{j=1}^{l} \beta_j \circ \eta  = \eta + \sum_{j=1}^{l}\gamma_j \circ h_{j-1}; \quad q = \upsilon \circ \eta + \sum_{l=1}^{L}\upsilon \circ \gamma_l \circ h_{l-1} 
\end{equation}
\citet{elhage2021mathematical} introduces the term \textit{residual stream} to describe $h_{l}$, a concept that can be traced back to~\citet{sepplong} and \citet{srivastava2015highway}.
\citet{veit2016residual} describe and study the unrolled structure of the final residual stream $h_L$, which reveals a number of paths from the input to the decoder that grows \textit{linearly} with the network depth.

\vspace{-3mm}
\paragraph{Linear residual networks.}

The presence of non-linearities at each block (and at the decoding module) prevents us from \textit{directly} expanding the input-to-output computation further.\footnote{One can still recover an exponential expansion of gradient paths when considering $\nabla q $, e.g. to analyze behavior during training, as \citet{veit2016residual} do. 
In this work, however, we solely focus on the forward dynamic of the network.} 
Linear residual networks, represented in \Cref{eq:lin-nets},
do not have this impediment. Indeed, if $\gamma_i(x)=A_i x$ for some $A_i\in\mathbb{R}^{d\times d}$, $\eta = E$ and $\gamma=\mathrm{id}$, we have that 
\begin{equation}
    \label{eq:lin-nets}
    \textstyle
    q = U(\sum_{S\in 2^{[L]}}  \prod_{j \in S} A_j) \, E = \sum_{S\in 2^{[L]}} q_S
\end{equation}
where $2^{[L]}$ is the power set of $[L]=\{1, \dots, L\}$ and the $q_S= U (\prod_{j\in S} A_j) E = U W_S E$, with $W_{\emptyset} = I$. \Cref{eq:lin-nets} writes (``expands'') the linear network into a combination of $2^L$ input-to-output paths $q_S:\mathcal{Z}\to\mathcal{Y}$, themselves linear functions. This enables a detailed analysis of each path's contributions (e.g. one may look at the norm of each $W_S$ as a measure of global path importance), roles, and interactions, as well as understanding global input-output relationships.
\begin{figure}
    \centering
    \includegraphics[width=0.68\linewidth]{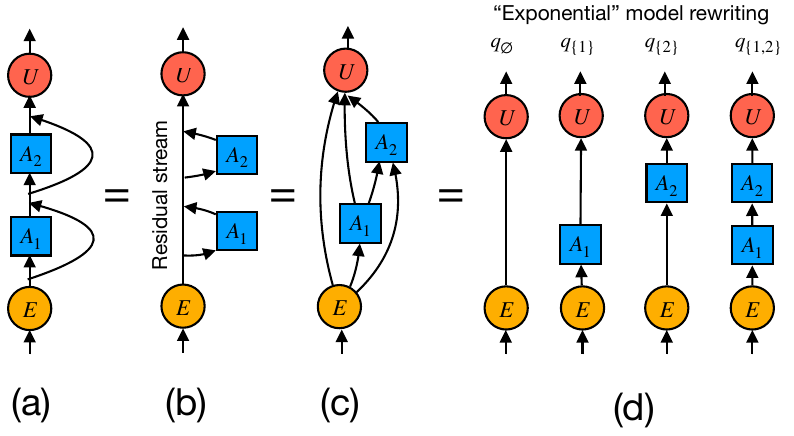}
    \caption{Various equivalent representations of a two-blocks linear residual network. In particular (b) highlights the residual stream of \cref{eq:unrolling}; (d) highlights the exponential rewriting of \Cref{eq:lin-nets}.}
    \label{fig:lin-nets}
    \vspace{-5mm}
\end{figure}

\section{Recursive  expansion of residual networks with jets}
\label{sec:re}
To tackle non-linearities and enable expansions in general residual networks similar to that of \Cref{eq:lin-nets}, we turn to jets~\citep{ehresmann1951}%
, which generalize Taylor expansions. In this section, we first introduce key concepts pertaining jets that are instrumental in developing our framework. Then we move to develop \decompose{}, the general algorithm for expanding residual nets into atomic input-output computational paths.

\vspace{-3mm}
\paragraph{Jet operators and their convex combinations} 

We recall that, for a function $f\in C^{k+1}(\rd, \rd)$ and $x, y \in \rd$, Taylor's theorem asserts that
\begin{equation}
    \label{eq:taylor-basics}
    \textstyle
    f(\vy) = f(\cx) + \textstyle \sum_{j=1}^k (j!)^{-1}\mathrm{D}^j f(\cx)(\vy-\cx)^{\otimes j} + O(\|\vy-\cx\|^{k+1}) 
\end{equation}
where $\cx,\vy$ are respectively the \centerptr{center} and \variateptr{variate}, $\mathrm{D}^j$ denotes the $j$-th differential,  $(y-x)^{\otimes j}$ denotes the $j$-fold tensor product, and $O(\|y-x\|^{k+1})$ denotes the class of functions that vanish at least as fast as a degree-$(k+1)$ polynomial $M\|y-x\|^{k+1}$ as $y\to x$ for some $M > 0$.
The $k$-th order jet operator of a function $f$ maps vectors to equivalence classes of degree-$k$ polynomial functions (we denote the resulting quotient space  by $P^k$ in the equation below, details in the appendix) as follows:
\begin{equation}
\textstyle
    \jet f: \rd \to \mathrm{P}^{k} \quad \quad
    \jet f (\cx) = f(\cx) + \textstyle \sum_{j=1}^k (j!)^{-1} \, \mathrm{D}^j f(\cx).
\end{equation}
Evaluating the jet at a variate $\vy\in\rd$ yields the truncated Taylor expansion $\jet f(\cx)(\vy)\in\rd$, that is, \Cref{eq:taylor-basics} without the ``$O$'' term.
The main advantage of working with jets rather than Taylor expansions is that we can work directly with functions rather than vectors.
We will make extensive use of the following lemma, of which the proof can be found in the appendix, along with further details about jets. 
\begin{lemma}[Convex combinations of jets] \label{lem:cc}
Let $f\in C^{\infty}(\rd, \rd)$,  $k\in\mathbb{N}, N\in\mathbb{N}^+$, $\{\centerptr{x_i}\}_{i\in[N]}$ be a set of \emph{jet centers},  $w\in\triangle^{N-1} \subset \mathbb{R}^N$ be a set of  \emph{jet weights},  and 
    $r=\max_i \{ w_i\| x_i - \sum_{j} x_j\| \}$. Then 
$$\textstyle
    \jet f\left(\centerptr{\sum_{i=1}^N x_i}\right)=
        \sum_{i=1}^N w_i \jet f(\centerptr{x_i}) + O(r^{k+1}).
        $$
\end{lemma}

\begin{remark}[Jet centers and variates as functions]
\label{rk:center-maps}
 We will often want to trace the computation of a jet back to the input space $\mathcal{Z}$. 
In such cases, we interpret the jet centers $\cx$'s and the variates $\vy$'s as functions of the original network input $z\in\mathcal{Z}$ onto $\rd$ or $\mathcal{Y}$.
Thus, we have that $\jet f(\cx)(\vy):\mathcal{Z}\to\rd$ (or $\mathcal{Y})$ which evaluates as follows: $\jet f(\cx)(\vy)(z) = \jet f(\centerptr{x(z)})(\variateptr{y(z)})$.
\end{remark}

\begin{figure}
    \centering
    \includegraphics[width=0.9\linewidth]{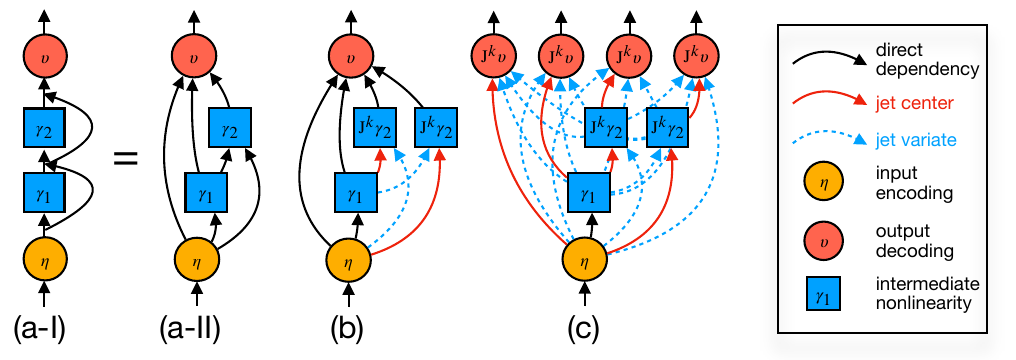}
    \caption{\begin{small}
        Representation of a two-blocks residual net (a, a-bis) and its exponential expansion steps (b, c).
    \end{small}}
    \label{fig:cartoons}
    \vspace{-3mm}
\end{figure}

\vspace{-3mm}
\paragraph{Exponential expansion of a two-blocks network.} 
Before introducing the main algorithm, we start with a minimal example of an expansion of a network with two residual blocks into four input-to-output paths. The network,  represented in \Cref{fig:cartoons} (a) and (a-bis), is given by:
\begin{equation}
    \label{eq:one-layer-net}
    q = \upsilon \circ h; 
    \quad h_2 =  \beta_2 \circ \beta_1 \circ \eta 
            = \eta 
              + \gamma_1 \circ \eta
              + \gamma_2 \circ (\eta + \gamma_1 \circ \eta )
\end{equation}
The final residual stream $h_2$ is a sum of three terms (input-to-hidden-space functions).
In a transformer network, $\gamma_1$ could represent a self-attention block and $\gamma_2$ an MLP block -- typically both transformations being input-normalized. 
Critically, the last term $\gamma_2 \circ (\eta + \gamma_1 \circ \eta )$ does not allow us to directly single out contributions that involve $\gamma_2$ and $\eta$ \textit{or} $ \gamma_1 \circ \eta$ alone.
To recover such paths, we can jet-expand $\beta_2$ and apply  \Cref{lem:cc} choosing as centers $x_\emptyset=\eta$ and $x_{\{1\}}=\gamma_1 \circ \eta$, obtaining:
\begin{equation}
\label{eq:2layers:1}
\begin{aligned}
    \jet \beta_2(x_\emptyset + x_{\{1\}}) = & w_1 \jet \beta_2(x_\emptyset) + w_2 \jet \beta_2(x_{\{1\}}) + O(r^{k+1}) \\
    = & x_\emptyset + x_{\{1\}} + w_1 \jet \gamma_2(x_\emptyset) + w_2 \jet \gamma_2(x_{\{1\}}) + O(r_{\beta_2}^{k+1}),
\end{aligned}
\end{equation}
where the last equality holds for $k\geq 1$. \footnote{For $k=0$ the weights apply also to the center terms since $\jet[0]\mathrm{id}(x_{\{1\}} + x_{\{2\}}) = w_1 x_{\{1\}} + w_2 x_{\{2\}} + O(r^1)$.}
This operation is represented in \Cref{fig:cartoons} (b).
These terms still do \textit{not} yield input-to-output paths, as in general $\gamma_3\neq \mathrm{id}$ (in transformer architecture this is typically a normalization operation, e.g. layer norm).
We can again proceed with a jet expansion, this time of the decoding module $\upsilon= U \, \gamma_3$. 
Continuing with our example, we apply \Cref{lem:cc} using as centers the outputs of the previous expansion, namely $x_\emptyset$, $x_{\{1\}}$, $x_{\{2\}}=w_1\jet \gamma_2(x_\emptyset)$ and 
$x_{\{1, 2\}} = w_2 \jet \gamma_2(x_{\{1\}})$, obtaining 
\begin{equation}
    \textstyle
    \jet \upsilon(x_\emptyset + x_{\{1\}} + x_{\{2\}} + x_{\{1,2\}}) = \sum_{S\in 2^{[2]}} \omega_1 U\,\jet \gamma_3(x_S) + O(r_{\upsilon}^{k+1})
\end{equation}
where $\omega\in \Delta^3$ is a vector of jet weights. 
With this operation, represented by \Cref{fig:cartoons} (c), we have obtained four input-to-output paths, mimicking the exponential rewriting of the linear case; cf. \Cref{eq:lin-nets}. 
For instance, the zeroth order ($k=0$) path that passes through the second non-linearity only, skipping the first, is given by the function 
$z \in\mathcal{Z} \to \omega_3 U \, \gamma_3(w_1\gamma_2(\eta(z)))\in \mathcal{Y}$. 
This example demonstrates the key principles of our approach: recursive expansion of the computational graph using jets, and the use of convex combinations to isolate specific paths. However, for deeper networks with many blocks, manually expanding each layer becomes impractical. To address this, we generalize this process into an algorithmic framework, which we develop next.

\begin{figure}[t]
    \centering
\begin{minipage}[t]{0.49\textwidth}
\begin{algorithm}[H]
\caption{\decompose($q$, $l$, $\mathcal{C}$, $k$)} \label{alg:jet-expand}
\begin{algorithmic}[1]
\Require Residual net $q$, block index $l\in [L]$; ~~ jet centers $\mathcal{C}=\{\cxi\}_{i\in[N]}$;
order $k\in\mathbb{N}$;
\Ensure $\xi$ is a set of (partial) jet paths with weights $w\in\triangle^{N-1}
$ and $\delta$ is 
a  reminder.
\State $\xi \gets 
\{w_i \jet {\gamma_{l+1}}(\cxi)\}_{i\in[N]}$
\If{$l < L$}
\State $\xi\gets \xi \cup \{w_i \jet \mathrm{id}(\cxi)\}_{i\in[N]} $
\State $\delta \gets h_{l+1} - \sum_{e\in\xi} e$ 
\EndIf
\State \textbf{else } ~$\delta \gets\gamma_{L+1} \circ h_L - \sum_{e\in\xi} e$

\end{algorithmic}
\end{algorithm}
\end{minipage}
\hfill
\begin{minipage}[t]{0.49\textwidth}
\begin{algorithm}[H]
\caption{\texttt{exp\_jet\_expansion}$(q, k)$} \label{alg:eje}
\begin{algorithmic}[1]
\Require Residual network $q$; order $k\in\mathbb{N}$;
\Ensure $\xi$ is a set of equally weighted input-to-output jet paths, $|\xi|=2^L$, and $\delta$ is 
a reminder.
\State $\xi\gets \{\eta, \gamma_1 \circ \eta \}$
\For{$l\in [L]$}
    \State $(\xi, \delta) \gets$ \decompose{}$(q, l, \xi, k)$
    \State $\xi \gets \{e(\cdot, 1/|\xi|)\}_{e\in \xi}$
\EndFor
\end{algorithmic}
\end{algorithm}
\end{minipage}
\vspace{-3mm}
\end{figure}

\vspace{-3mm}
\paragraph{The \texttt{jet-expand} algorithm.}

\Cref{alg:jet-expand} presents the key operation of the framework.
The algorithm applies \Cref{lem:cc} to a residual transformation or to the decoding non-linearity for a given (user-defined)
set of centers $\mathcal{C}$.
It yields a set of expanded polynomial terms $\xi$, which can be seen as a set-valued function $
\xi:\mathcal{Z} \times \triangle^{N-1} \to \mathcal{E}$,
where $\mathcal{E}$ is an appropriate power set of functions, 
and a non-linear remainder $\delta: \mathcal{Z} \times \triangle^{N-1} \to \rd$. 
The remainder encompasses both the residuals stemming from \Cref{eq:taylor-basics} and \Cref{lem:cc}.
As we showed above, centers can be the outputs of previous expansions, enabling the propagation of the expansion through the entire network and effectively 'unrolling' the computation graph into distinct paths.
Importantly, once we apply the algorithm for $l=L$ we obtain a way to \textit{rewrite the computational graph} of $q$ as a sum of expanded terms (input-to-output paths), which we call \textit{expansion}, and a non-linear remainder.
Indeed, if $(\xi_L, \delta_L)=$\decompose{}$(q, L, \mathcal{C}, k)$ for some $\mathcal{C}$ and $k$, the following class of functional equivalences holds:
\begin{equation}
    \label{eq:jet-rewrite}
    \textstyle
    q = \sum_{e \in \xi_L} U
    \, e(\cdot, w) + \delta_L(\cdot, w) \qquad \text{for } w\in \triangle^{N-1}.
\end{equation}
The runtime of \Cref{alg:jet-expand} is negligible as it operates at the level of the computational graph. 
Evaluating $\xi$ (and $\delta$) at any $z
\in\mathcal{Z}$, instead, incurs a runtime complexity of $O(|\mathcal{C}|(F+ kB))$ where $F$ and $B$ are the costs of a forward and a backward evaluation of $q$, respectively. 
In practice runtime can be reduced by storing computation \citep{griewank2008evaluating, bettencourt2019taylormode}.
In next section we discuss how particular instantiations of our framework encompass previous studies and let us seamlessly define novel objects of interest such as $n$-gram statistics of LLMs. 
Before that, we conclude the section with two remarks regarding  remainders and jet weights.
\begin{remark}[Non-vanishing remainders]
\label{rmk:non-vanishing}
In general, we cannot expect reminders to vanish (as $k$ grows). Indeed, even if the convergence radius of the Taylor series is infinite, the arguments of residuals introduced by applications of \Cref{lem:cc} do not vanish.
If $q$ is a linear residual network, however, $\delta =0$ for $k\geq 1$, showing that \Cref{alg:jet-expand} recovers (after reorganizing terms) the rewrite of \Cref{eq:lin-nets} for every choice of $w$. \footnote{Other special cases include expansions where each center set is a singleton and the convergence radius of the expanded non-linearities is infinite.}  
Hence, in light of \Cref{eq:jet-rewrite}, jet expansions should be seen as ways to rewrite computational graphs rather than approximations; in experiments we show however how $\delta$'s can be small and the cosine similarity between expansion and original network logits can be close to 1; see \Cref{fig:lenses} (bottom).
\end{remark}
\begin{remark}[Jet weights optimization]
    So far we glossed over the role of the jet weights $w$'s. In principle, these can be fixed, e.g. $w_i=1/N$. 
    However, jet weights can also be optimized to minimize the remainder at any given $z$, e.g. after projecting it into the logit space. 
    Interesting, this can be done cheaply as $\| U \delta_L(z, w) \|^2= \| \gamma_L(h_L(z)) -  \sum_{e\in \xi_L}§ e(z, w)\|^2_{U^T U}$, which amounts to the squared distance between the expansion and the original residual stream in the representation space $\rd$ with the metric induced by the unembedding matrix.
\end{remark}

\section{Notable expansions and their implications}
\label{sec:notable}

We introduce some particular expansions as application of the introduced $\decompose{}$ algorithm, setting the stage for the numerical case studies of the next section.

\vspace{-3mm}
\paragraph{(Super)exponential expansion.} \Cref{alg:eje} generalizes the exponential expansion we performed onto the two-blocks network in \Cref{sec:re}, using uniform jet weights. 
One can interpret the algorithm as performing a ``maximal'' expansion (when remaining at the grain of the blocks) which yields $2^L$ input-to-output paths.
In fact, for $k\geq 1$, we can further isolate each degree of the expanded terms into separate input-to-output paths that highlight interactions among various blocks. 
This further refinement, which we will focus on in future work, may suggests that residual networks may in fact behave as super-exponential ensembles of (shallower) functions.

\vspace{-3mm}
\paragraph{Jet lenses and logit lens.}
The logit lens \citep{nostalgebraist2021interpreting, geva2021FFN,geva2022FFN,merullo2023language,belrose2023eliciting} is an interpretability method that consists in applying the decoder to intermediate representations as follows:
\begin{equation*}
\textstyle
\mathrm{LogitLens}_{l}(z) = U \gamma(h_l(z)) = \jet[0]\upsilon (h_l(z))(h_L(z)).
\end{equation*}
The logit lens, aimed at highglighting the iterative refinement of the prediction across blocks,
is related to early exiting (or early decoding) in the context of conditional computation \citep[see e.g.][]{panda2016,elbayad2020depth,geva2022FFN}. 
It is immediate to verify that $\mathrm{LogitLens}_{l}$ 
is equivalent to the expansion yielded by \decompose{}$(q, L, \{h_l\}, 0)$. 
This suggests two generalizations, which we dub \textit{iterative} and \textit{joint} jet lenses, respectively.
The iterative jet lens is a direct extension of the logit lenses with higher order jets: \decompose{}$(q, L, \{h_l\}, k)$. The joint jet lenses are expansions obtained  through \decompose{}$(q, L, \{\gamma_l \circ h_{l-1}\}_{l\in[L]}, k)$ that are aimed at highlighting the residual contributions of each block non-linearity, rather than the iterative refinement of the residual stream.

\vspace{-3mm}
\paragraph{Jet bi-grams and skip-$n$-grams statistics.} 
We consider transformer-based large language models with alternating self-attentions and MLPs, which are particular instances of residual nets. \footnote{We disregard positional embeddings for simplicity and leave their study to future work.}
Our framework allows us to directly extract $n$-gram statistics from an existing LLM without any probing datasets.
Concretely, we can systematically evaluate relevant jet paths (for small $n$'s) on the entire input space, usually the vocabulary and its Cartesian products, independently from individual contexts.
For example, bi-grams statistics related to $\mathbb{P}_q(z_2| z_1, \dots)$ can be computed by evaluating bi-gram paths, which we can obtain by expanding the LLM with \Cref{alg:eje} and filtering out all paths that involve self-attention modules. 
Specifically in our case studies~(\Cref{sec:cases}), we focus on encoding-decoding bi-gram path, obtainable via expanding the LLM with \decompose{}$(q, L, \{\eta\}, k=0)$, and the bi-gram paths involving up to one MLP module, which can also be obtained via applying \Cref{alg:jet-expand} twice.
We can obtain skip-$n$-gram statistics relating to $\mathbb{P}_q(z_n|z_{n-1}, \dots, z_{n-2}, \dots, z_{1}, \dots)$, where dots indicate any number of interceding tokens, by evaluating jet paths with self-attentions (the fewer self-attentions, the lower the $n$) and isolated single query-key products. 
Such jet $n$-gram statistics offer a \emph{data-free} tool to sketch LLMs via casting them into (symbolic) $n$-gram databases.
Thus they allows us to perform symbolic model diffing between \textit{any} two models that share a common vocabulary, 
as opposed to take differences in the parameter space, harder to interpret and only possible for same-architecture models.  

\section{Interpreting LLMs with jet expansions}
\label{sec:cases}
Our framework provides users with freedom in terms of choosing the computational paths they wish to focus on. 
Jet expansions support studies across various levels, including model-level global analysis (jet $n$-grams), component-level analysis (jet paths), and example-level analysis (jet lens).
We experiment with several popular open-sourced large language models families: \textit{GPT-2}~\citep{gpt2}, \textit{GPT-Neo}~\cite{gpt-neo}, \textit{Llama}~\citep{llama,touvron2023llama,codellama} and \textit{OLMo}~\citep{groeneveld2024olmo}, showcasing the generality of the algorithm.
Our main experiments run on $128$ CPU servers with $1$ TB memory, while jet lens experiment run on a single laptop.

\subsection{Analyzing LLM inner working}
\label{case:inner_working}
LLMs are notorious for their lack of interpretability due to their inherent model complexity and size, made worse by the usual opaque training process and unknown training data. Understanding their inner working contributes to calibrating trust for users to use them appropriately. We showcase how jet expansion along user-selected computational paths (jet paths) can help us discover and locate learned associations akin to studies in mechanistic interpretability~\citet{templeton2024scaling}.
\vspace{-3mm}
\paragraph{Jet lenses.}

\begin{figure}
    \centering
    \includegraphics[width=1.\linewidth]{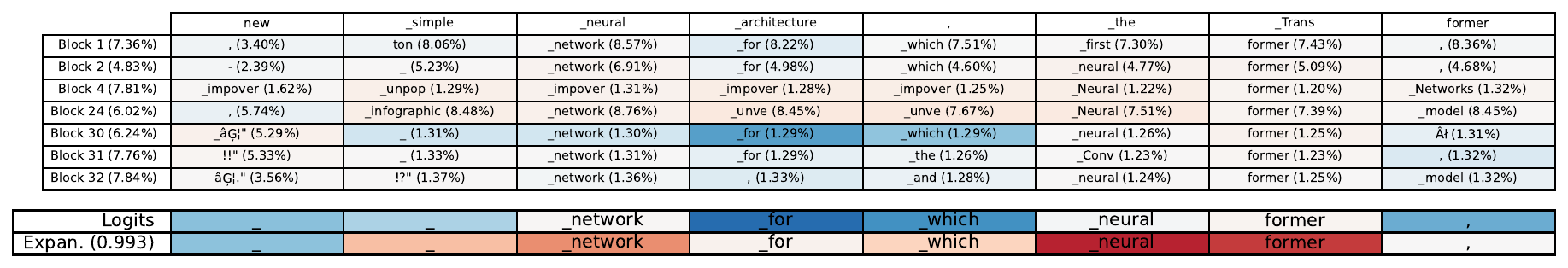}
    \includegraphics[width=1.\linewidth]{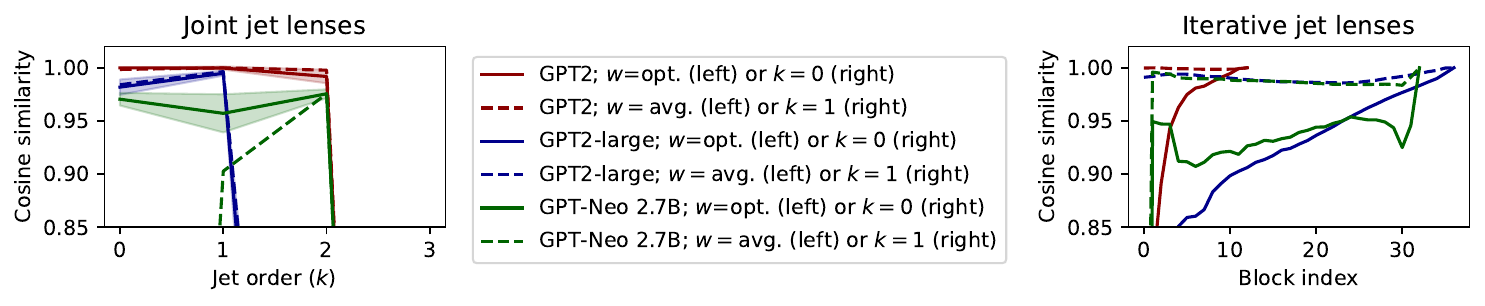}
    \caption{
    \begin{small}
    (\textbf{Top}) example of a joint jet lens on \textit{GPT-Neo $2.7$B} with $k=1$, visualizing the seven blocks with highest average jet weights after optimization. 
    Each table cell indicates the most likely token of the jet path related to each block non-linearity.  Optimized jet weight are in brackets. 
    We used a diverging blue-to-red color map tracking logit scores, centered around zero.
    The bottom table shows the model logits and the expansion logits, with cosine similarity in brackets; in this case, all top-$1$ tokens perfectly coincide.
    (\textbf{Bottom}) plots of average cosine similarities between original and jet logits of joint (\textbf{left}) and iterative (\textbf{right}) lenses.
    \end{small}
    }
    \label{fig:lenses}
    \vspace{-5mm}
\end{figure}
We use jet lenses introduced in \Cref{sec:notable} to analyze LLM's mechanism when processing individual examples. 
\Cref{fig:lenses} (top) visualize a joint jet lens for GPT-Neo 2.7B \citep{gpt-neo} (other examples can be found in \Cref{sec:apx-lenses}). 
Here, a block contains one self-attention and one MLP module. 
All table cells depict top-$1$ tokens for the corresponding path, following conventions from prior work~\citep{belrose2023eliciting}.
We observe that the joint jet lens captures the synergy among different blocks, as the model prediction is decomposed into several jet paths.
Our preliminary analysis supports recent work on super-position~\citep{elhage2022superposition} and neuron polysemy~\citep{bricken2023monosemanticity}, suggesting that interactions among components may have ensemble effects, which can broadly vary across model families. 
In this sense, the jet lenses with $k>0$ may serve as tools to systematically discover such synergic behaviors.
We also find that higher-orders ($k>0$) help iterative lenses deliver more meaningful interpretations than the logit lens ($k=0$) for \textit{GPT-Neo-$2.7$B} (see  \Cref{fig:jet_k0_neo,fig:jet_k1_neo,fig:jet_k2_neo}). This is potentially due to their capability to trace indirect impacts of early layers on the final logits, which were otherwise missing under logit lens. 
Our findings are consistent with ~\cite{nostalgebraist2021extension,cancedda2024spectral} where naive implementations of logit lens are shown to fail on \textit{GPT-Neo} model family.
\Cref{fig:lenses} (bottom) present mean cosine similarities of joint and iterative jet lenses for various \textit{GPT} models and orders, averaged over $100$ example sentences. 
The similarities are high and close to $1$ for various $k$, showing however different behavior across model families and sizes. This indicates jet expansions highly correlate with model outputs, potentially providing faithful interpretations.

\vspace{-3mm}
\paragraph{Jet paths of individual components.}
 By examining the representative jet bi-grams that are captured by each MLP path, we find some MLPs that perform special linguistic functions.
 For example, in \textit{OLMo-$7$B}, the jet path which passes through the $3$rd MLP  promotes the addition of the ``\texttt{-ing}'' suffixes to the current token. 
 Similar MLPs with certain linguistic functions are listed in \Cref{tab:ffn_function}. Note that the relationship between functions and components are not necessarily one-to-one mappings. 
 Particularly we find that the paths through multiple MLPs might work together to complete one linguistic function e.g. 
 MLP $6$ and MLP $18$ in \textit{Llama-$2$-$7$B} can add ``\texttt{-ing}'' suffix. One MLP might also do multiple linguistic jobs e.g. MLP $1$ in OLMo 7B adding ``\texttt{-ly}'' and ``\texttt{-▁else}'' suffixes. This echos work on circuit discovery~\citep{arthur2023circuit,ferrando2024circuit} and superposition~\citep{elhage2022superposition}, where the role of each component cannot be easily dissected and multiple components collaborate to fulfill a function. 
 \Cref{tab:attention_head_roles} reports a role identification study on attention heads in the first self-attention of \textit{OLMo-$7$B} using jet tri-grams. 
 Specifically, we find heads associated with math and programming, e.g. head $1$ on Math/Latex; heads promoting digits and dash composition into dates, e.g. head $25$; and heads constituting phrase templates, e.g. head $15$ managing a ``for $x$ purposes'',  where $x$ is a placeholder.
 To verify the roles we revealed, we further perform preliminary intervention experiments where we ablate MLPs or attention heads and compute  variations in model logits.
 After the interventions, the logits drop consistently in all cases, suggesting our jet $n$-grams indeed can help identify certain roles for selected components.
 Varying impact on logit differences is likely due to overdetermination \citep{mueller2024missed} and our partial selection of jet paths (e.g. for tri-grams we only selected encoding-attention-decoding paths, excluding any MLP).

\begin{table}[t]
\caption{MLPs in \textit{OLMo-$7$B} and \textit{Llama-$2$-$7$B} performing certain linguistic functions based on jet bi-grams extracted from the corresponding jet paths. 
}
\label{tab:ffn_function}
\centering
\resizebox{0.9\textwidth}{!}{
\begin{tabular}{llllll|llll}
\toprule
 & \multicolumn{5}{c}{\textit{OLMo-$7$B}} & \multicolumn{4}{c}{\textit{Llama-$2$-$7$B}} \\
\midrule
\textbf{MLP Index} & $1$ & $3$ & $9$ & $17$ & $19$ & $6$ &  $7$ & $18$ & $19$ \\
\midrule
\textbf{Role} & \texttt{-ly, -▁else} & \texttt{-ing} & \texttt{-'t} & \texttt{-▁than} & \texttt{-s} & \texttt{-ing} & \texttt{-es} & \texttt{-ing,-ity} & \texttt{-ly}  \\
\textbf{$\Delta$ logit after intervention} & $-4.19, -3.35$ & $-0.58$ & $-9.73$ & $-4.26$ & $-7.42$ & $-14.61$ & $-3.55$ & $-9.69, -11.93$ & $-9.14$ \\
\bottomrule
\end{tabular}
}
\end{table}
\begin{table}[t]
\centering
\caption{Several attention heads in the first residual block of \textit{OLMo-$7$B} and their roles identified with jet tri-grams extracted from corresponding jet paths. We also include an example tri-gram captured by each head.}
\resizebox{\linewidth}{!}{%
\begin{tabular}{ccccc}
\toprule
\textbf{Head Index} & $2$                  & $16$                               & $26$                & $30$                               \\ 
\midrule
\textbf{Role}                 & Math/LaTeX          & ``for \dots purposes''       & date composition    & ``into account/consideration \dots''   \\ 
\textbf{Example 3-gram}              & \texttt{(▁Lemma, ▁let, ▁s)} & \texttt{(▁for, ▁use, ▁purposes)} & \texttt{(20, 23, ▁-)} & \texttt{(▁into, ▁account, ▁possible)}      \\ 
\textbf{$\Delta$logit after intervention}           & $-0.1570$             & $-0.0019$                           & $-0.0093$             & $-0.0001$ \\
\bottomrule
\end{tabular}%
}
\label{tab:attention_head_roles}
\end{table}

\subsection{Analyzing pretraining dynamics}
\label{case:pre}
Pretraining an LLM is usually extremely resource intensive.
Therefore it is crucial to monitor the progress of a pretraining run to prevent wasting of time and compute.
In this section, we show how jet bi-grams can serve as an effective signaling tool to trace the pretraining dynamics, providing insights about the model's maturity. Such signals are especially useful to understand what happens with the model when the pretraining loss shows marginal improvements and fails to reflect the changes inside the model.

\vspace{-3mm}
\paragraph{Identifying the top bi-grams.} To assess the model's progression, we extracted jet bi-grams from \textit{OLMo-7B} model checkpoints across $555$K pretraining steps. \Cref{tab:diff_steps} presents a summary of the top $10$ jet bi-grams at different stages of training. Due to space reason, we only show the top $10$ jet bi-grams every $100$K steps.  Initially, the network exhibits nonsensical jet bi-grams, such as ``\texttt{ICUirling}''.
As training advances, it gradually learns more meaningful combinations, like ``\texttt{at least}''. This process of acquiring sensible bi-grams stabilizes around step $200$K, indicating that the model is reaching a level of maturity where the top $10$ bi-grams capture common  meaning.

\begin{figure}[t]
    \centering
    \begin{subfigure}{0.485\linewidth}
        \centering
        \includegraphics[width=\linewidth]{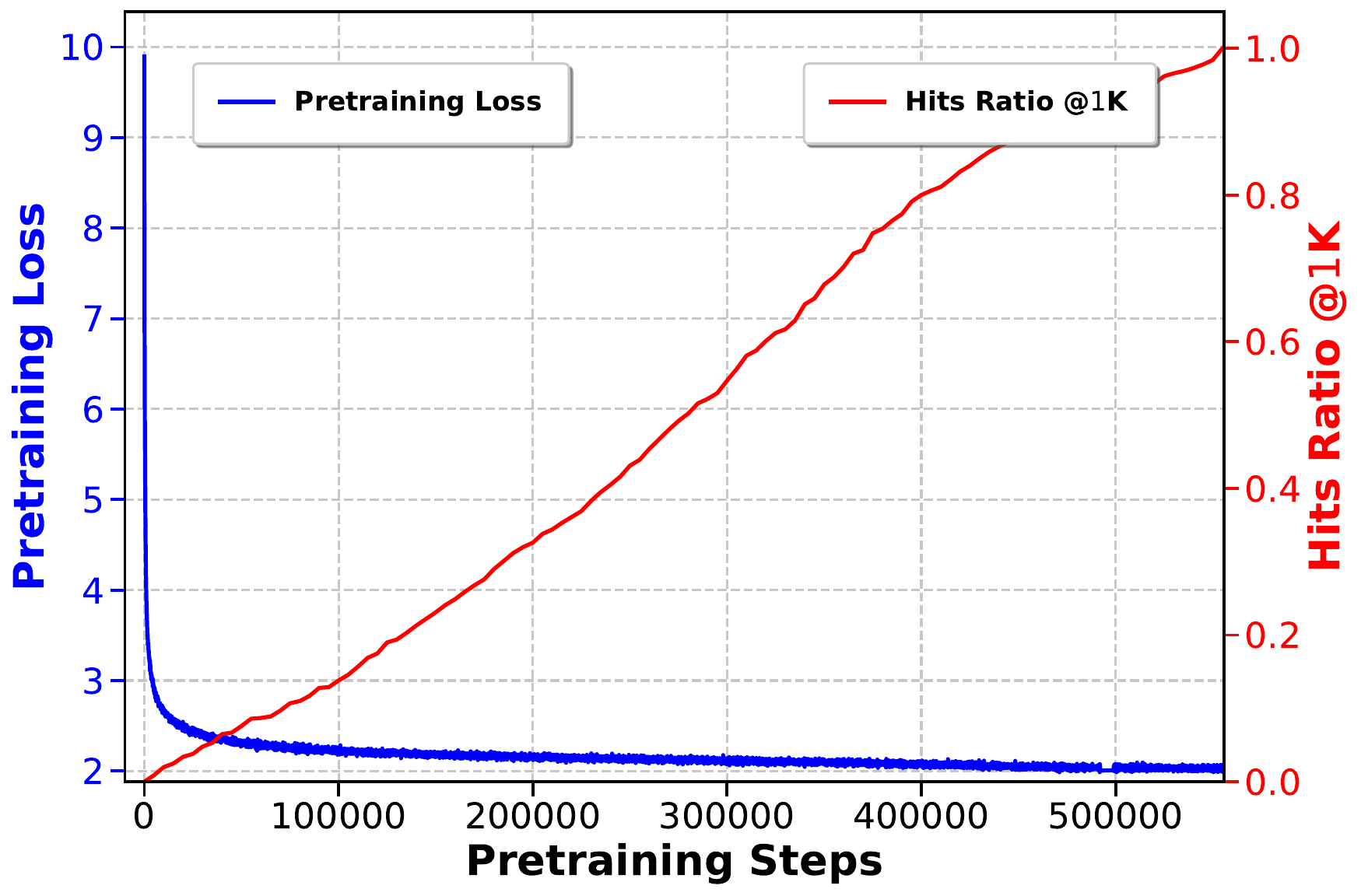}
        \caption{Top $1$K jet bi-gram hit ratios w.r.t. the final step.}
        \label{fig:evolution}
    \end{subfigure}
    \hfill
    \begin{subfigure}{0.485\linewidth}
        \centering
        \includegraphics[width=\linewidth]{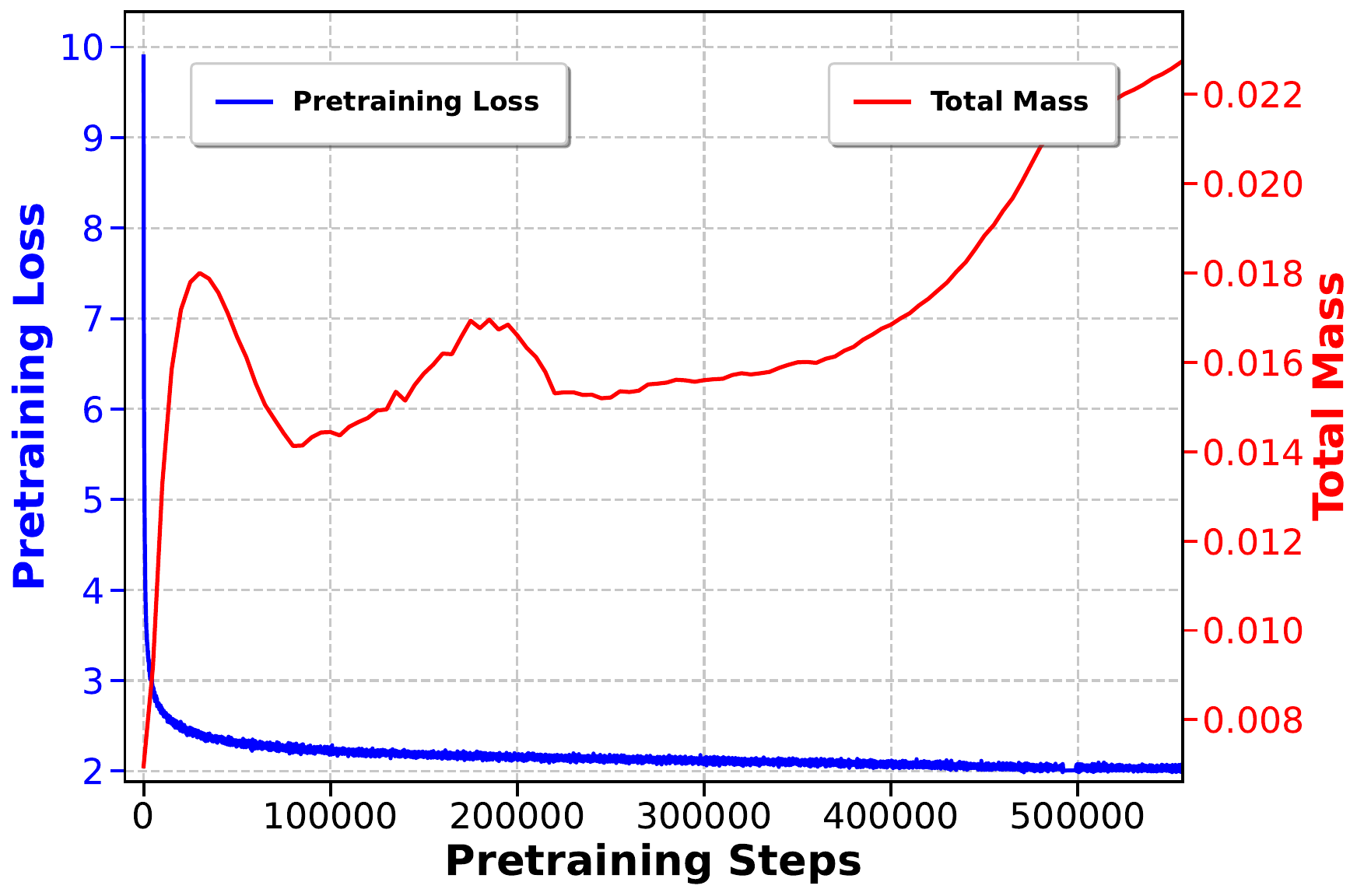}
        \caption{Top $1$K jet bi-gram mass w.r.t. empirical data.}
        \label{fig:evolution_oracle}
    \end{subfigure}
    \vspace{-2mm}
        \caption{Analysis of \textit{OLMo-$7$B}'s pretraining dynamics via measuring its jet bi-gram progression.}
        \vspace{-4mm}
\end{figure}

\vspace{-3mm}
\paragraph{Analyzing bi-grams learning speed.} To evaluate the learning speed of these jet bi-grams, we consider the jet bi-grams at the final training step ($555$K) as the ground-truth bi-grams. 
We then chart the hit ratios of these ground-truth bi-grams at each pretraining step, as illustrated in \Cref{fig:evolution}. Interestingly, even though the pretraining loss (the blue curve) shows only minor improvements after the initial $50$K steps, the model’s acquisition of effective bi-grams continues to progress in a steady, consistent manner. This observation aligns with known phenomena in neural network training, such as double-descent and grokking, which highlight the model's ability to improve generalization capabilities even when the loss appears to stagnate~\citep{zhang2021understanding,power2022grokking}.
In addition, \Cref{fig:evolution_oracle} characterizes the total pseudo-joint probability mass of top $1$K bi-grams from empirical data~\citep{liu2024infini}.
We derive a pseudo-joint jet bi-gram probability using statistical uni-grams from \citep{liu2024infini}.
We observe that the model gradually accumulates probability mass that aligns with the real corpus data distribution.

\begin{figure}[t]
\centering
\includegraphics[width=0.8\linewidth]{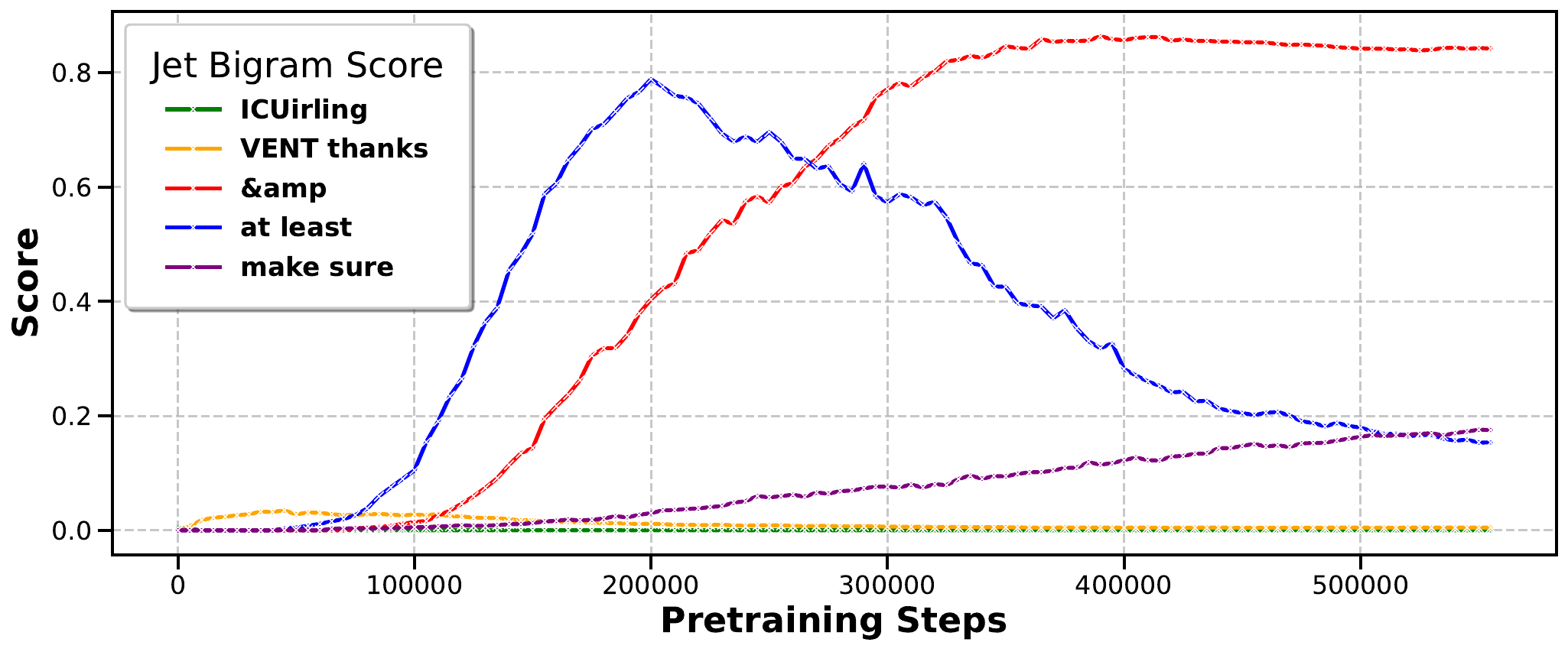} 
\caption{Visualization of  \textit{OLMo-$7$B}'s promotion and suppression dynamics of jet bi-grams scores.}
\label{fig:promotion_suppression}
\vspace{-3mm}
\end{figure}

\vspace{-3mm}
\paragraph{Learning schemes for different bi-grams.} To understand if there are any differences between the learning schemes of different bi-grams, we can trace the progression of the jet bi-gram scores for selected bi-grams. \Cref{fig:promotion_suppression} provides a visual comparison of how different bi-grams are promoted or suppressed during the pretraining process. 
The different slopes and levels of the lines indicate varying rates of learning for the respective bi-grams.
We observe that, the model first acquires random bi-grams due to random parameter initialization. These random bi-grams, like ``\texttt{ICUirling}'' and ``\texttt{VENT thanks}'', are quickly suppressed in the early steps and never regain  high scores. 
In contrast, one-to-many bi-grams like ``\texttt{at least}'' are first promoted to very high scores but then get suppressed perhaps due to the model seeing more of the scope of the token ``at''.  One-to-one bi-grams like ``\texttt{\&amp}'' (HTML code) are gradually promoted and stabilize. 
Many-to-many bi-grams like ``\texttt{make sure}'' takes the most time to learn and the scores are still increasing even at the end of pretraining. 
Our findings suggest that the training process effectively promotes certain ``good'' bi-grams, but at different paces, where they might be suppressed later depending on their occurrences and linguistic nature. These insights could inform future training strategies, such as targeted training on more relevant bi-grams or adjusting the training data to improve the pretraining speed.

\subsection{Analyzing fine-tuning effect}
\label{case:fte}
Fine-tuning is an important phase where the raw pretrained LLMs are guided to perform particular tasks. 
We would like to understand how the model inner knowledge changes during fine-tuning processes.
While parameter diffing can be a straightforward solution, jet n-grams provides an alternative approach, where the diffs are human readable and directly reflect the change of knowledge retained by the LLMs.
Such insights would allow us to better decide the mixture of data for fine-tuning, and the number of steps for fine-tuning, which are currently a mix of heuristics and trial-and-error.

\vspace{-3mm}
\paragraph{Code fine-tuning promotes coding-relevant bi-grams.} We analyze the changes due to code fine-tuning via \emph{diffing} jet bi-grams extracted from \textit{Llama-2-7B} and its fine-tuned versions, \textit{Codellama-7B} and \textit{Codellama-Python-7B}. As highlighted in \Cref{tab:diff_finetuning} with orange coloring, the jet bi-gram diff reveals coding-relevant keywords, such as ``\texttt{**kwargs}'', ``\texttt{getters}'' and ``\texttt{Assertion}'', suggesting jet bi-gram can be a tool for verifying if fine-tuning is effective in acquiring relevant knowledge. 

\begin{table}[t]
\centering
\caption{Toxicity indexes for \textit{Llama-$2$-$7$B} and \textit{Llama-$2$-$7$B-chat} using different methods: \textit{ToxiGen}, jet bi-grams, and \textit{RealToxicityPrompts} challenge prompting. Higher numbers indicate higher toxicity scores on the corresponding benchmarks and higher toxic knowledge possession for jet bi-grams.}
\resizebox{\linewidth}{!}{%
\begin{tabular}{lcccccc}
\toprule
 & \multicolumn{1}{c}{ToxiGen Score} & \multicolumn{1}{c}{Jet Bi-grams} & \multicolumn{4}{c}{RTP Challenging Prompts} \\
\cmidrule(lr){2-2} \cmidrule(lr){3-3} \cmidrule(lr){4-7}
 & \small{\cite{hartvigsen-etal-2022-toxigen}} & Mass of ``toxic'' bi-grams & No & Very mild & Medium & Hard \\
\midrule
\textit{Llama-$2$-$7$B} & $21.25$ & $0.03445$ & $38\%$ & $49\%$ & $64\%$ & $88\%$ \\
\textit{Llama-$2$-$7$B-chat} & $0.0$ & $0.03377$ & $23\%$ & $35\%$ & $64\%$ & $84\%$ \\
\bottomrule
\end{tabular}%
}
\vspace{-5mm}
\label{tab:merged_toxicity}
\end{table}

\vspace{-3mm}
\paragraph{Does RLHF fine-tuning remove toxicity?} We compare the original pretrained model, \textit{Llama-2-7B}, with its RLHF version, \textit{Llama-2-7B-Chat}. RLHF alignment~\citep{bai2022training} is widely believed to detoxify LLMs, as indicated by the \textit{ToxiGen} scores~\citep{hartvigsen-etal-2022-toxigen}. However, it remains easy to prompt LLMs to bypass this alignment and produce toxic content. 
In \Cref{tab:merged_toxicity}, we demonstrate this with dataset-based toxicity scores on a subset of challenging prompts in the \textit{RealToxicityPrompts} (RTP) dataset~\citep{gehman-etal-2020-realtoxicityprompts}: the gap in toxicity potential between the two models \textit{narrows} as we prepend to RTP prompts increasingly ``explicit'' (short) context.
Specifically, for hard context, \textit{Llama-2-7B-Chat} shows an $84\%$ probability of producing toxic content, close to that of \textit{Llama-2-7B}.
This suggests that the RLHF model is not completely detoxified but rather hides the toxicity knowledge from the ``surface'', which however can be easily  triggered by specific contexts.
To quantify the toxicity knowledge embedded in these models, we use jet bi-gram probability scores and calculate the cumulative conditional probability mass for a set of ``toxic'' bi-grams, which are combinations of tokens associated with toxic meanings from a predefined list of keywords.
Interestingly, we observe a small change in mass from $0.03445$ to $0.03377$ after RLHF.
Thus, although the \textit{ToxiGen} score may suggest that the model has been effectively detoxified, the jet bi-gram mass reflects retention of toxic knowledge after RLHF, aligning with the scores obtained by  introducing medium or hard explicit context and computing a toxicity score (via a second scorer model, \citep{Detoxify}) on   \textit{RealToxicityPrompts} dataset~\citep{gehman-etal-2020-realtoxicityprompts}.
This showcases a potential application of jet bi-grams in constructing \emph{data-free} indices that reveal embedded knowledge, offering complimentary views beyond traditional data-driven benchmark evaluations.

\section{Related work}
\label{sec:rw}
\paragraph{Interpreting transformers.}
There has been much effort in interpreting the inner computations of transformer models. In particular, \emph{mechanistic interpretability} \cite{ferrando2024primer}, focuses on reverse-engineering such computations by identifying, clustering and labelling model behavior \citep{shah2024decomposing, meng2022locating, bricken2023monosemanticity} in human understandable terms and attributing them with certain model components, e.g., MLPs \cite{geva2021FFN, geva2022FFN}, or typical ``circuits" \citep{arthur2023circuit, ferrando2024circuit}.
Authors discussed limitations of currents approaches to MI. 
For example, \citet{templeton2024scaling} found it generally hard to conclude neuron-level intepretabilities, compared with feature representations; 
while \citet{bolukbasi2021interpretability, goldowsky2023localizing} points out that conclusions drawn are generally limited to the chosen data distribution. 
As our approach focuses on manipulating functions, it does not require extra datasets that are used for probe fitting in methods such as \citet{belrose2023eliciting} nor sampling, as needed in \citep{arthur2023circuit,ferrando2024circuit,voita-etal-2024-neurons}.
On a high level, allowing taking any portion of compute out of the original transformer, jet expansions abstract and generalize previous characterizations on the computational paths \citep{veit2016residual, elhage2021mathematical}, where non-linear components with significant roles, e.g. layernorm and MLPs, are either ignored or over-simplified for the ease of analysis.
Additionally, zero ablations (or knock out) \citep{olsson2022context} and direct logits attributions \citep{wang2022interpretability} are linked to particular instantiations of zeroth order jet expansions.

\vspace{-3mm}
\paragraph{$n$-gram models.} The early applications of $n$-gram models in languages dates back to \citep{shannon1948mathematical}, where $n$-grams modeled the statistics of English. The $n$-gram based approaches have been an important baseline in language processing, e.g., general language modelling \citep{goodman2001bit} with applications like machine translation \citep{brants2007large}. There have been regained interests on combining $n$-gram with neural network model-based approaches \citep[e.g.][]{liu2024infini}. 
Several recent works have explored the relationships between LLMs and $n$-gram language models, 
such as analyzing the representational capacity of transformers to simulate $n$-gram LMs \citep{svete2024transformers} and measuring agreement between LLM predictions and $n$-gram rulesets~\citep{nguyen2024understanding}. 
\vspace{-3mm}
\paragraph{Taylor expansion and jets}
Taylor expansions are popular tools in analyzing learning  behaviours \citep{jastrzkebski2017residual}, notably linearization ($k=1$). For example, \citet{belrose2024neural} applied Taylor expansion on the loss function to demonstrate the learning preference of neural network models. \citet{xu2022information} introduced a second-order Taylor expansion over the data distribution to interpret optimal features. 
The generalized jet notions was introduced in machine learning in the context automatic differentiation tools by \citet{bettencourt2019taylormode}, and is an experimental feature in Jax \citep{jax2018github}, but has been studied before \citep[see e.g.][]{griewank2008evaluating}.

\section{Conclusion and discussion}

We introduced \textit{jet expansion}, a novel framework for expanding the computational graphs of  neural networks. 
The method, which we specialize in this paper to deep residual nets, can be used to disentangle contributions of user-selected computational paths from the overall  graph. 
Complementary to other dataset-dependent methods in MI, our method enables various dataset-free global interpretability studies, such as mapping computation to linguistic roles. 
We have validated jet expansions in terms of cosine similarity against model outputs and through interventional experiments (\Cref{case:inner_working}). 
We applied our data-free method to transformer LMs, showing how we can sketch the original model with input-output probability databases, extracting LM bi-and-tri-gram statistics.

\vspace{-3mm}
\paragraph{Limitations.}
Although rooted in Taylor series theory, expansions obtained via our frameworks do not (seek to) approximate the input function in any strict sense. 
Rather, our framework is amed at  facilitating interpretation of model behavior: we can use jet expansion \textit{to rewrite} an input computational graph as a sum of ``interpretable'' polynomial terms and a (computable) remainder.
How large is a reminder and how expansions align with  model outputs remains at the moment an empirical question, implying that the jet order and weight optimization routines should generally be considered as hyperparameters of the method.
Furthermore, expansions are not unique (but higher order expansions "contain" lower order one).
We leave a deeper  investigation of these aspects to future work.
From a runtime standpoint,
we note that even though graph manipulation is almost immediate, systematic evaluation of jet paths may be time consuming~(especially for $k\gg0$ and when optimizing jet weights).
If the input space is large, one may need to resort to sub-sampling or search heuristics. 
Finally, we limited our study of $n$-gram expansions of LMs to bi-and-tri-grams,  unearthing compelling behaviors. This leaves the study of longer-context expansions to future work. 

\vspace{-3mm}
\paragraph{Implications and future work.}
Our work opens up several research directions.
From a theoretical standpoint, we will extend the expansion procedure to cover finer granularities, e.g. at neuron~(subspace) levels; incorporate established attribution methods such as the Shapley value \citep{shapley1953value}, including recent extensions to deal with probabilistic models \citep{franceschi2024explaining}; develop concepts of (approximate) equivalence classes over models leveraging the jet spaces, which, in turn, may further ground the model diffing procedure sketched in our case studies. 
Furthermore, we will take inspiration from the numerous tools in linear algebra to provide further depth into the analysis, deepening the link to linear residual structures and 
establishing relations with Markov chains and hidden Markov models, recently employed e.g. by  \citet{zhang2023tractable} for constrained (structured) decoding. 
We plan to investigate the implication of the super-exponential number of paths in the residual networks depth unearthed by \Cref{alg:eje}.  
From an applications standpoint, besides studying jet $n$-grams for $n>3$, we envision several fruitful  applications in safety and transparency, such as developing ``search features'' to systematically detect unwanted associations, or leaked private content. 
Although our experiments are primarily observational, we speculate that \decompose{} may also become an useful tool to guide interventions, supplementing other techniques like causal tracing \citep{meng2022locating} and  path patching \citep{goldowsky2023localizing}.

\bibliography{main}
\bibliographystyle{iclr2025_conference}

\newpage
\appendix

\section{Additional details on jets}
A jet of a function represents an equivalence class. We thus can perform algebraic operations among functional equivalence classes using jet algebra stated below.

\begin{proposition}[Jet algebra]
\label{prop:jet-algebra}
Let $f,g\in C^{\infty}(\rd, \rd)$ and $k\in\mathbb{N}^+$. Then,
\begin{itemize}
\itemsep0em
    \item[(i)] $\jet (a f + bg)(\cx) = a \, \jet(f)(\cx) + b \, \jet(g)(\cx)$, for $a, b\in\mathbb{R}$ (linearity);
    \item[(ii)] $\jet f(\cx) \circ g \in \jet f(\cx)
    $ and $\jet f(\cx) \circ g(y)=\jet f(\cx)(\variateptr{g(y)})$ (jet after endomorphisms);
    \item[(iii)] 
    $g \circ \jet f(\cx) = 
    \{ g\circ u  \, : \, u \in \jet f(x) \}$ (endomorphism after jet); 
    \item[(iv)] $\jet (f \circ g)(\cx) = \jet f(\centerptr{g(x)}) \circ \jet g(\cx)$ (composition of jets);
\end{itemize}
\end{proposition}
Properties \textit{(i)}-\textit{(iii)} follow directly from the definition; \textit{(iv)} is a consequence of the chain rule and truncation.

\paragraph{Proof of \Cref{lem:cc}}
\label{sec:proof_of_combination}
Take $y \in \mathbb{R}^d$, $N \geq 1$, $x_i\in \mathbb{R}^d$ for $i\in[N]$,  $w\in \triangle^{N-1}$ and an order $k\geq 0$.
Since $w$ belongs to the simplex $\triangle^{N-1}$, we have $\sum_{i=1}^{N} w_i = 1$. Multiplying $f(y)$ on both hands, we obtain   
$$
f(y) = \sum_{i=1}^N w_i f(y) = \sum_{i=1}^N w_i \left[
f(x_i) + \sum_{s=1}^k \mathrm{D}^s f(x_i)(y - x_i)^{\otimes s} + O(\| y - x_i \|^{k + 1} ) 
\right]
$$
$$
= \sum_{i=1}^N w_i \mathrm{J}^k f(\centerptr{x_i})(\vy) + O(w_i\| y - x_i \|^{k + 1} ), 
$$
by applying \Cref{eq:taylor-basics} (Taylor expansion) and the definition of jet with each $x_i$ as the center.
At the same time, we can expand $f(y)$ with $\sum_{i=1}^N x_i$ as the center 
$$
f(y) = \mathrm{J}^k f(\centerptr{\sum_{i=1}^N x_i})(\vy) + O(\|y - \sum x_i\|^{k+1} ). 
$$
Now let us take $y=\sum_{i=1}^N x_i$ and observe that $O(\|y - \sum x_i\|^{k+1} ) = 0$ and 
$O(w_i\| y - x_i \|^{k + 1} ) = O(w_i\| x_i - \sum_j x_j \|^{k + 1} )$. Finally we observe that the class of functions in the last $O$ are dominated by the class of function in $O(r^{k+1})$ where $r=\max_i \{ w_i\| x_i - \sum_{j} x_j\| \}$ is the maximum remainder. 
This concludes the proof. 

As a side note, jet weights would not need to form convex combinations, but rather linear combinations $\sum_i w_i = 1$. 
However, restricting to convex combinations has two major advantages:
\begin{itemize}
    \item optimizing over a convex set guarantees the existence of maxima and minima (Weierstrass theorem) and uniqueness of minima if we are optimizing a strictly convex loss as in general is the case for expansions that only affect the decoder module.
    \item weights within the probability simplex have a clearer interpretation for interpretability purposes.
\end{itemize}

\section{Additional Tables for Jet Bi-grams}

See \Cref{tab:diff_steps} and \Cref{tab:diff_finetuning}.

\newcolumntype{L}{>{\rmfamily}l}
\newcolumntype{C}{>{\rmfamily}c}
\begin{table}
    \caption{Bi-gram evolution across pretraining steps for OLMo 7B. Each column represents a distinct step, while each row corresponds to a different rank. The table entries are the bi-grams at each step for each rank. The number of tokens seen in association with the pretraining steps is also annotated. The model gradually picks up meaningful bi-grams after starting from senseless bi-grams (due to random initialization).}
    \label{tab:diff_steps}
    \centering
    {\ttfamily
\resizebox{0.8\textwidth}{!}{%
\begin{tabular}{Cllllll}
\toprule
    \multirow{2}{*}{\rm{\bf Rank}} &  \textrm{0K} {\rm[{\bf \#steps}]}    &  \textrm{100K}  &  \textrm{200K}  &  \textrm{300K}  &  \textrm{400K}  &  \textrm{555K} \\
    
      &  \textrm{0B} {\rm[{\bf\#tokens}]}   &  \textrm{442B}  &  \textrm{885B}  &  \textrm{1327B}  &  \textrm{1769B}  &  \textrm{2455B}  \\
\midrule
0 &  immortal & ’s &  at least &  \&amp &  \&amp &  \&amp \\
1 &  ICUirling &  at least & ’s &  at least &  its own &  its own \\
2 & ords architect &  its own &  \&amp &  its own &  their own &  their own \\
3 & yaml Adam & okerly &  your own &  your own &  at least &  his own \\
4 & 231 next & VENT thanks &  its own &  their own &  your own &  make sure \\
5 & clonal\begin{CJK}{UTF8}{gbsn}条\end{CJK} & iums & iums &  more than &  his own &  your own \\
6 &  Charg@\{ &  you're &  you're &  can't &  2nd &  2nd \\
7 &  avoir careless & Everything v &  2nd &  his own &  more than &  at least \\
8 &  HOLD worsening & erna already &  you guys &  2nd &  make sure &  more than \\
9 &  Horse dismant & 'my &  more than &  make sure &  can't & iums \\
\bottomrule
\end{tabular}
}
}
\end{table}

\begin{table}
    \caption{The bi-grams before and after coding-finetuning. For space reason, we only show the bi-grams at every 50 ranks among the top 1000 bi-grams. We highlight the bi-grams that are relevant to coding, such as ``**kwargs'' a keyword in python programming. This demonstrate that our method has the capability to extract representative bi-grams that reflect fine-tuning quality.}
    \label{tab:diff_finetuning}
\centering
\resizebox{0.8\textwidth}{!}{%
{    \ttfamily \footnotesize
\begin{tabular}{llll}
\toprule
 \multirow{1}{*}{{\bf Rank}} & \multicolumn{1}{c}{\rm \textbf{LLAMA2-7B}} & \multicolumn{1}{c}{\rm \textbf{CodeLLAMA-7B}} & \multicolumn{1}{c}{\rm \textbf{CodeLLAMA-Python-7B}} \\
\midrule
\textrm{0} & (▁more, ▁than) & (▁like, wise) & (▁like, wise) \\
\textrm{50} & (▁Now, here) & (▁just, ification) & (▁Like, wise) \\
\textrm{100} & (▁system, atically) & (▁in, ▁case) & (▁all, udes) \\
\textrm{150} & (▁all, erg) & \textcolor{orange}{(▁get, ters)} & (▁no, isy) \\
\textrm{200} & (▁on, ions) & (któber, s) & \textcolor{orange}{(output, ted)} \\
\textrm{300} & (▁other, world) & (▁all, ud) & \textcolor{orange}{(Object, ive)} \\
\textrm{350} & (▁Just, ified) & (gebiet, s) & \textcolor{orange}{(▁as, cii)} \\
\textrm{400} & (▁trust, ees) & (▁Protest, s) & (▁can, nab) \\
\textrm{450} & (▁at, he) & \textcolor{orange}{(▁deploy, ment)} & (▁transport, ation) \\
\textrm{500} & (▁book, mark) & (Class, room) & \textcolor{orange}{(Tag, ging)} \\
\textrm{550} & (▁from, \begin{CJK}{UTF8}{gbsn}而\end{CJK}) & (▁access, ory) & (▁personal, ized) \\
\textrm{600} & (▁WHEN, ever) & (▁In, variant) & (▁excess, ive) \\
\textrm{650} & (▁where, about) & (▁I, ▁am) & (▁Add, itional) \\
\textrm{700} & (ag, ged) & (add, itionally) & \textcolor{orange}{(▁**, kwargs)} \\
\textrm{750} & (▁he, he) & \textcolor{orange}{(▁invalid, ate)} & (name, plates) \\
\textrm{800} & (▁all, anto) & \textcolor{orange}{(div, ision)} & (▁select, ive) \\
\textrm{850} & (▁Tom, orrow) & \textcolor{orange}{(▁process, ors)} & \textcolor{orange}{(▁Assert, ions)} \\
\textrm{900} & (▁for, ays) & (▁Program, me) & (blog, ger) \\
\textrm{950} & (▁Bach, elor) & \textcolor{orange}{(▁set, up)} & (▁can, cellation) \\
\bottomrule 
\end{tabular}
}}
\end{table}

\section{Additional plots of jet lenses}
\label{sec:apx-lenses}

See plots below, referring to the main paper for details. 
Note that for iterative lenses the last block coincides with the model logits for all $k$ by design.
We omit the iterative lens for GPT2-large for $k=2$ due to low cosine similarity.

\begin{figure}
    \centering
    \includegraphics[scale=0.45]{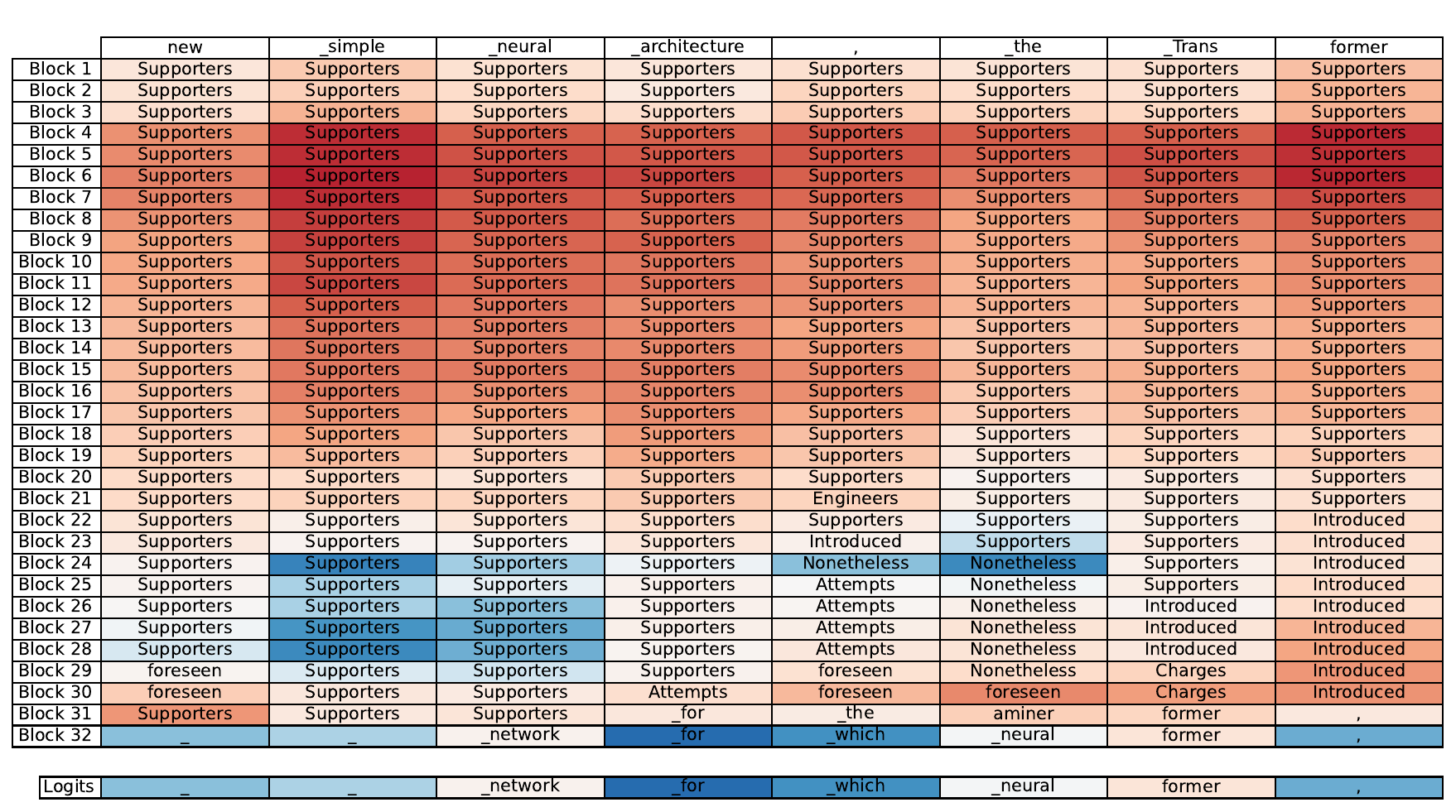}
    \caption{Iterative jet lens ($k=0$), equivalent to logit lens\citep{nostalgebraist2021interpreting}, applied over GPT-Neo-$2.7$B with the input sentence ``new simple neural architecture, the Transformer''.}
    \label{fig:jet_k0_neo}
\end{figure}

\begin{figure}
    \centering
    \includegraphics[scale=0.45]{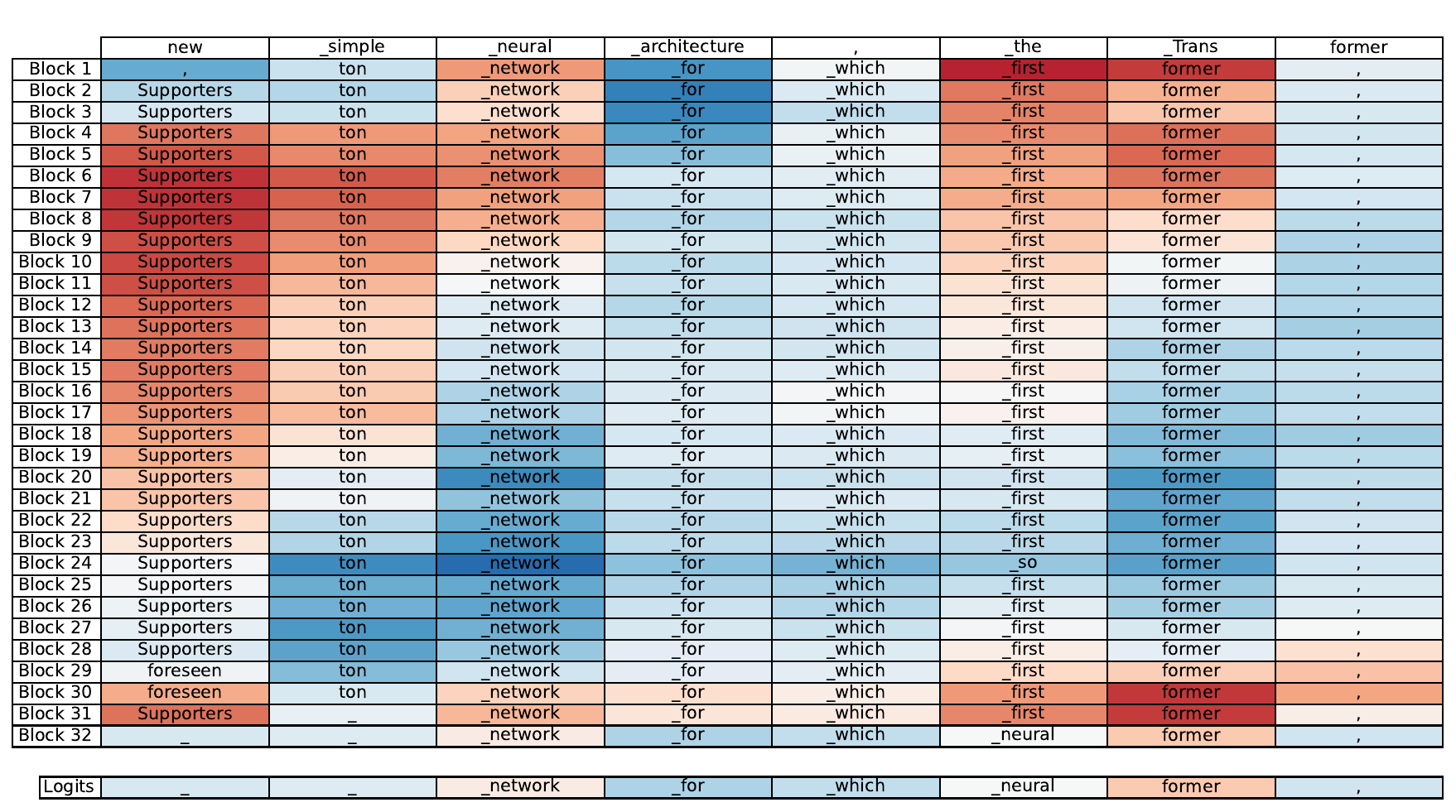}
    \caption{Iterative jet lens ($k=1$), applied over GPT-Neo-$2.7$B with the input sentence ``new simple neural architecture, the Transformer''}
    \label{fig:jet_k1_neo}
\end{figure}

\begin{figure}
    \centering
    \includegraphics[scale=0.45]{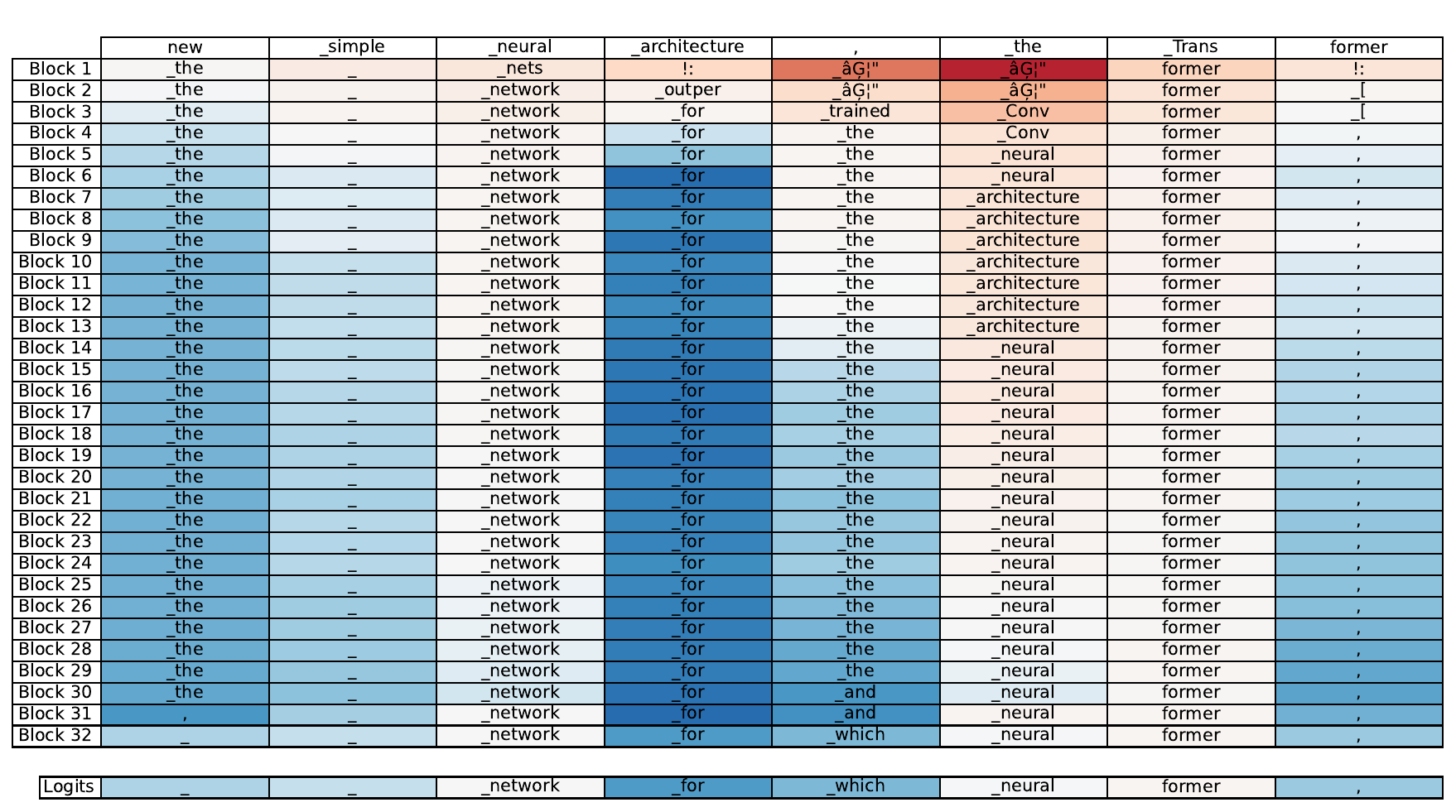}
    \caption{Iterative jet lens ($k=2$), applied over GPT-Neo-$2.7$B with the input sentence ``new simple neural architecture, the Transformer''}
    \label{fig:jet_k2_neo}
\end{figure}

\begin{figure}
    \centering
    \includegraphics[scale=0.45]{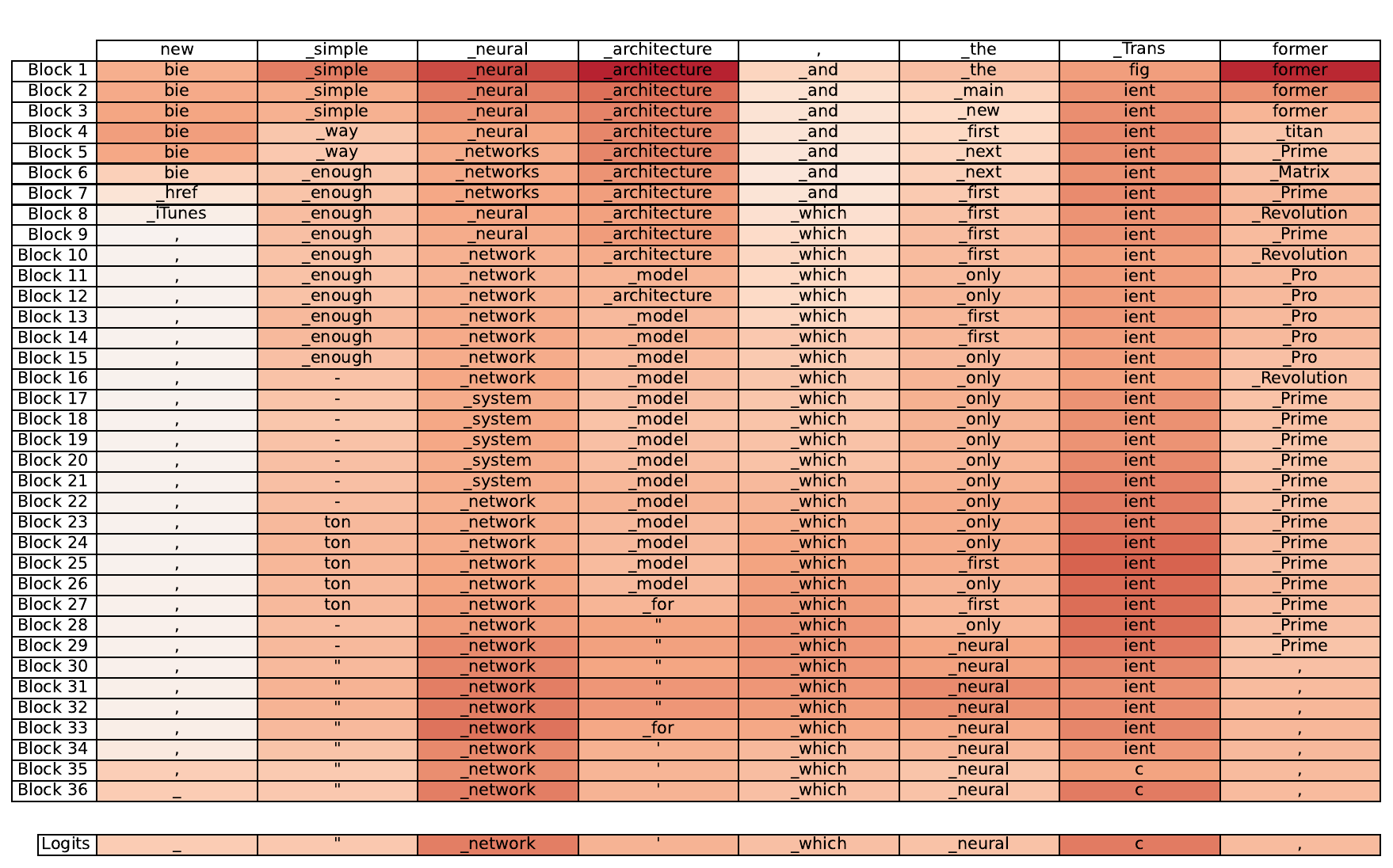}
    \caption{Iterative jet lens ($k=0$), equivalent to logit lens\citep{nostalgebraist2021interpreting}, applied over GPT-2-large with the input sentence ``new simple neural architecture, the Transformer''.}
    \label{fig:jet_k0_gpt2_large}
\end{figure}

\begin{figure}
    \centering
    \includegraphics[scale=0.45]{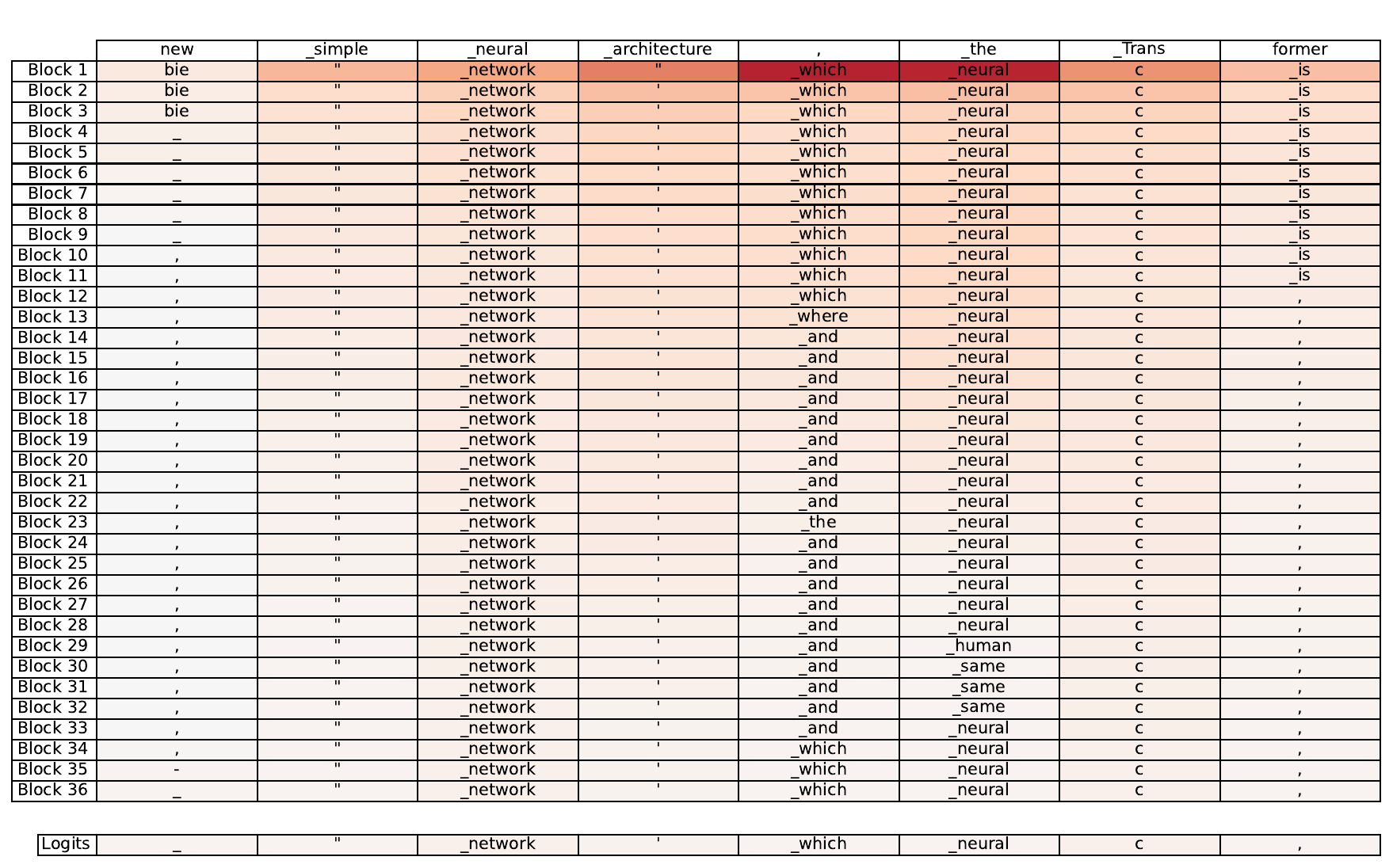}
    \caption{Iterative jet lens ($k=1$), applied over GPT-2-large with the input sentence ``new simple neural architecture, the Transformer''}
    \label{fig:jet_k1_gpt2_large}
\end{figure}

\begin{figure}
    \centering
    \includegraphics[scale=0.45]{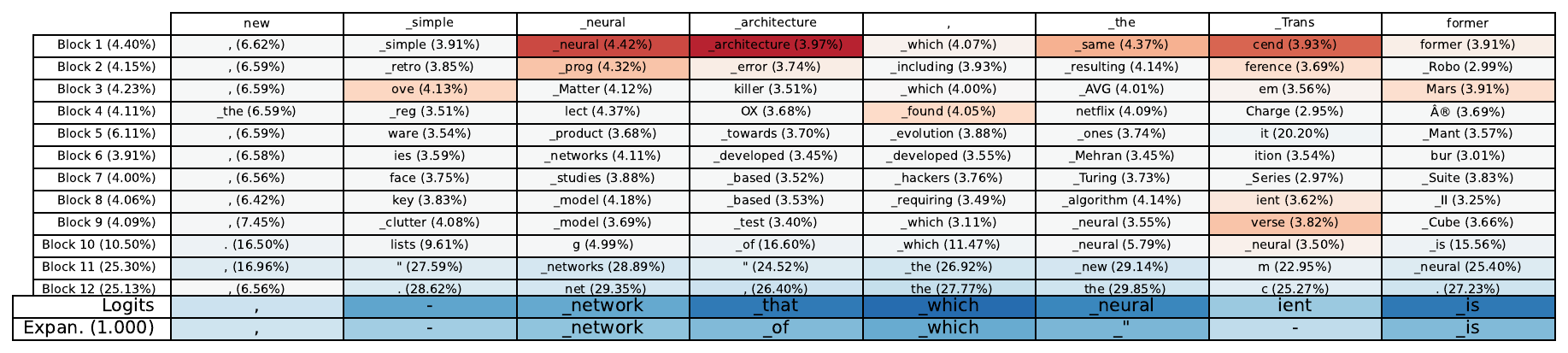}
    \caption{Joint jet lens with learnable weightings~($k=0$), applied over GPT2 with the input sentence ``new simple neural architecture, the Transformer''}
    \label{fig:jet_k0_gptbase_weighting}
\end{figure}

\begin{figure}
    \centering
    \includegraphics[scale=0.45]{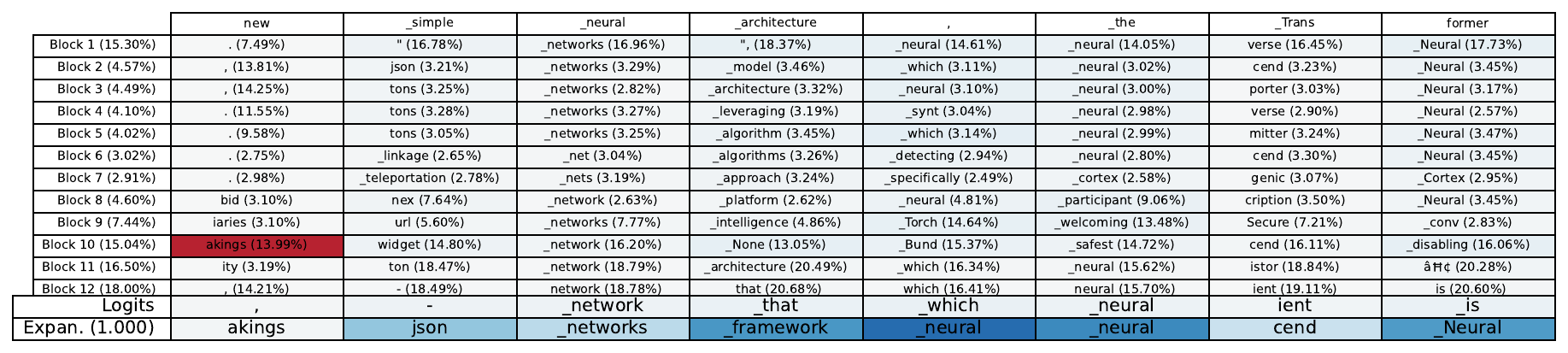}
    \caption{Joint jet lens with learnable weightings~($k=1$), applied over GPT2 with the input sentence ``new simple neural architecture, the Transformer''}
    \label{fig:jet_k1_gptbase_weighting}
\end{figure}

\begin{figure}
    \centering
    \includegraphics[scale=0.45]{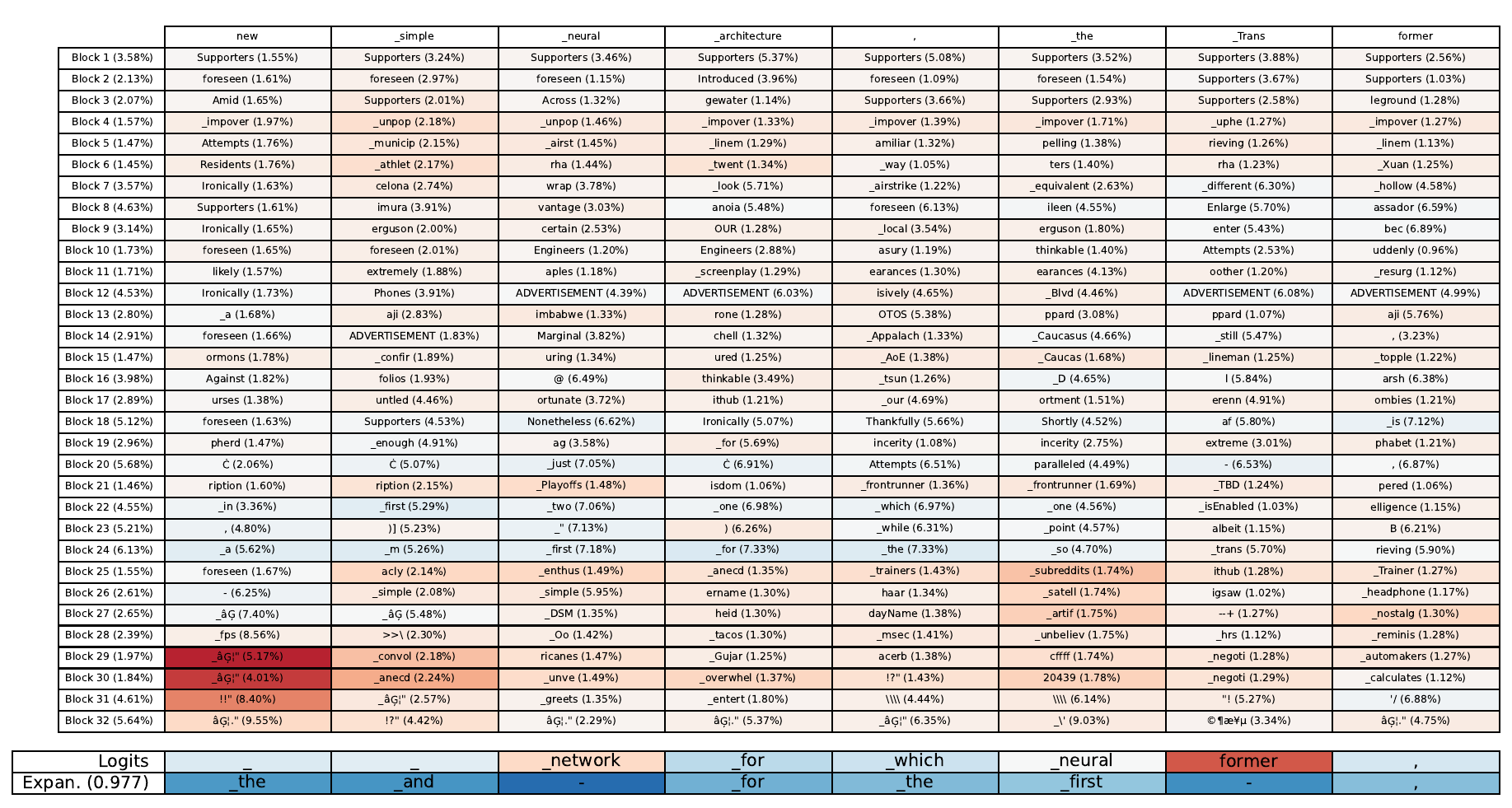}
    \caption{Joint jet lens with learnable weightings~($k=0$), applied over GPT-Neo-$2.7$B with the input sentence ``new simple neural architecture, the Transformer''}
    \label{fig:jet_k0_neo_weighting}
\end{figure}

\begin{figure}
    \centering
    \includegraphics[scale=0.45]{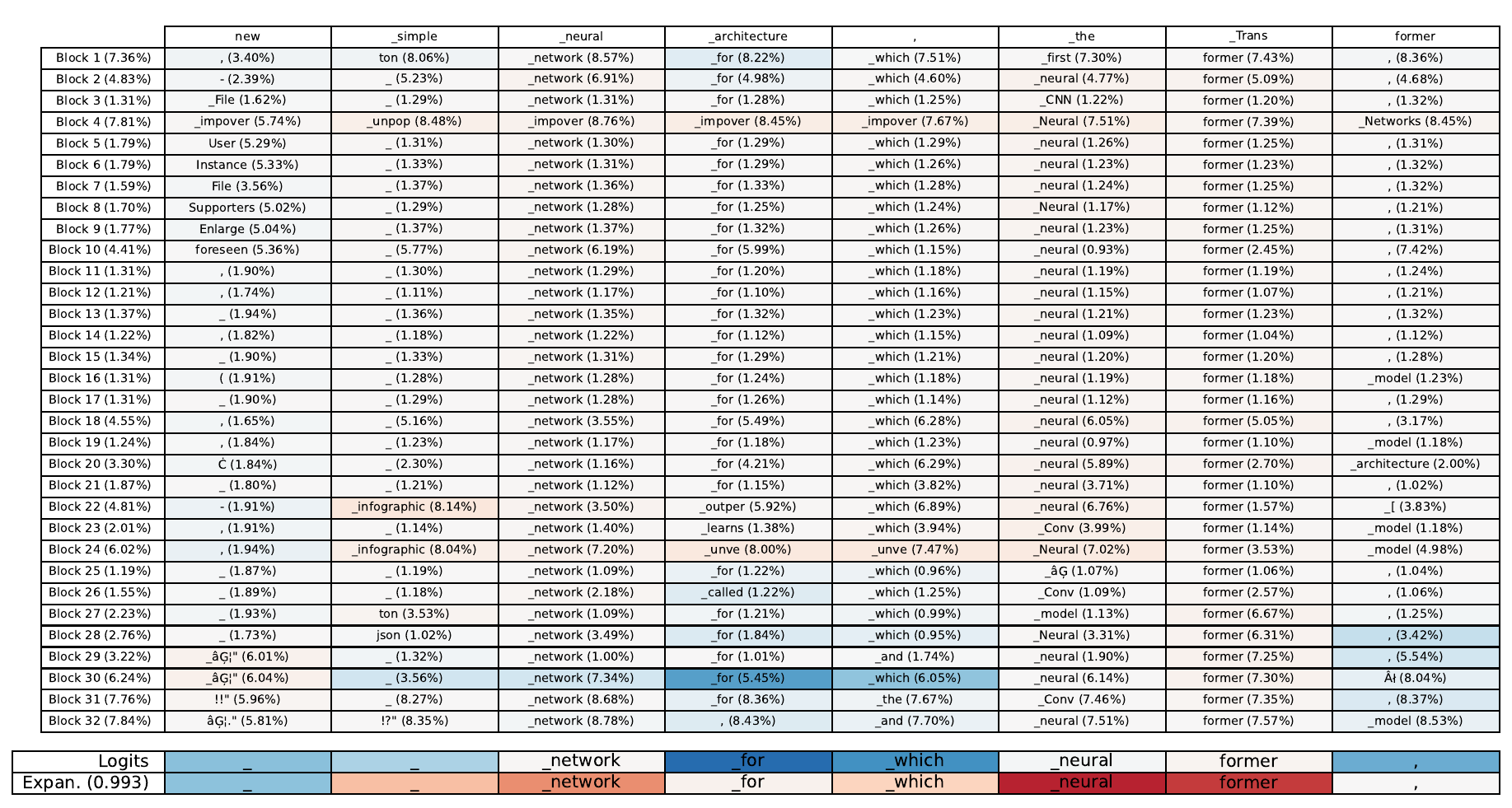}
    \caption{Joint jet lens with learnable weightings~($k=1$), applied over GPT-Neo-$2.7$B with the input sentence ``new simple neural architecture, the Transformer''}
    \label{fig:jet_k1_neo_weighting}
\end{figure}

\begin{figure}
    \centering
    \includegraphics[scale=0.45]{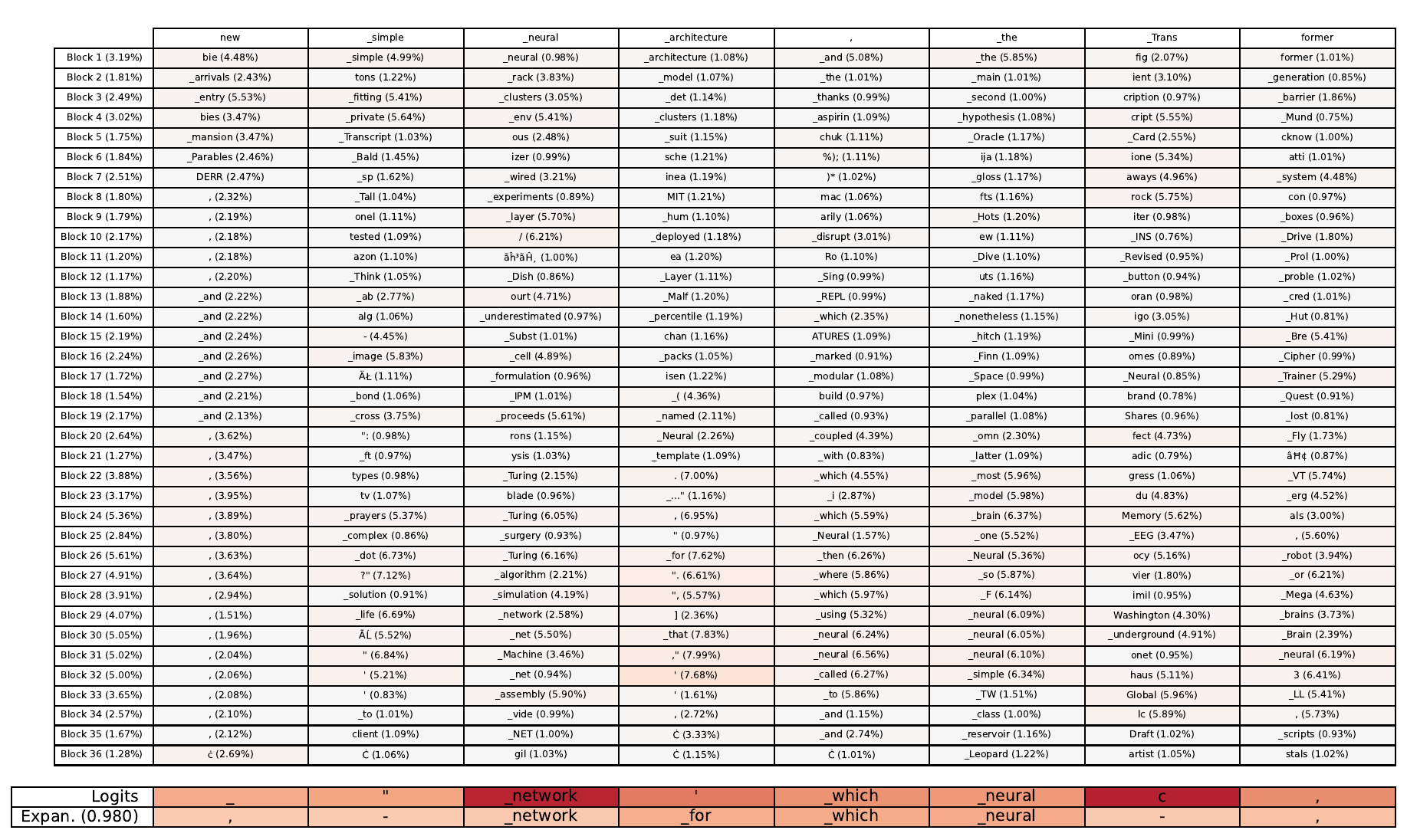}
    \caption{Joint jet lens with learnable weightings~($k=0$), applied over GPT-2-large with the input sentence ``new simple neural architecture, the Transformer''}
    \label{fig:jet_k0_gpt2_large_weighting}
\end{figure}

\begin{figure}
    \centering
    \includegraphics[scale=0.45]{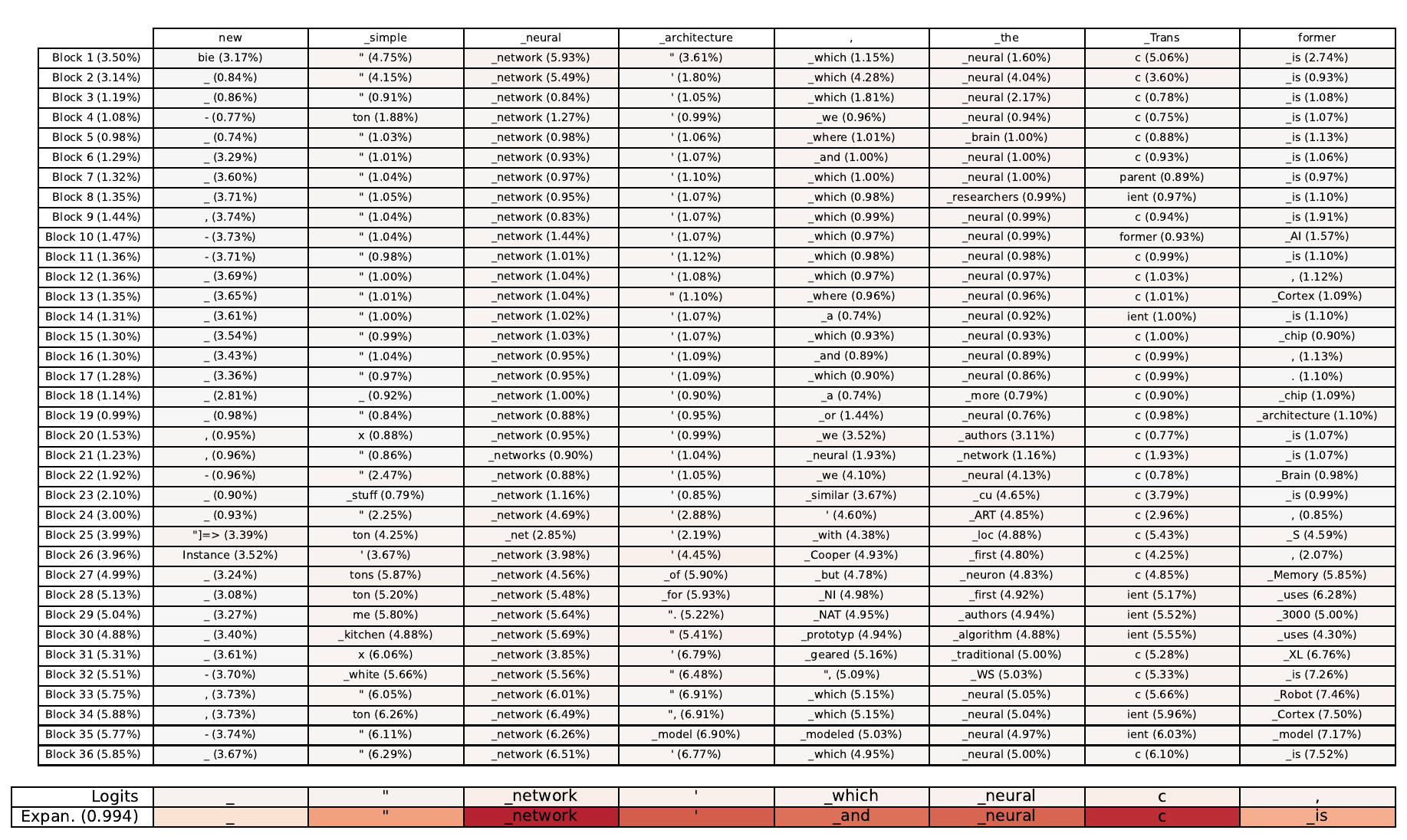}
    \caption{Joint jet lens with learnable weightings~($k=1$), applied over GPT-2-large with the input sentence ``new simple neural architecture, the Transformer''}
    \label{fig:jet_k1_gpt2_large_weighting}
\end{figure}

\section{Runtime}

\begin{figure}
    \centering
    \includegraphics[width=0.8\linewidth]{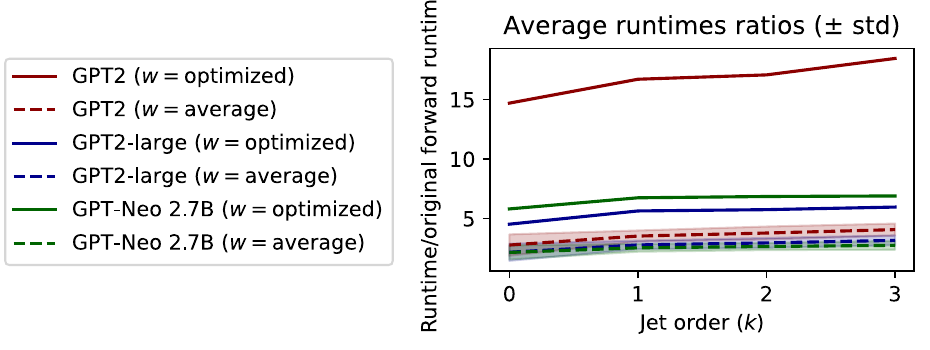}
    \caption{Empirical runtime of evaluations of jet expansions originating form the joint jet lenses as a ratio of the evaluation of the input model.}
    \label{fig:runtime}
\end{figure}

We report in \Cref{fig:runtime} a plot of the runtime for evaluating expansions originating from the joint jet lenses of \Cref{case:inner_working} as a ratio of the input model evaluation (forward pass), for both the uniform and the optimized jet weight setup, for different jet orders.

\end{document}